\documentclass{article}
\pdfoutput=1

 \usepackage[nonatbib, preprint]{nipstemplate/neurips_2022}
\usepackage[utf8]{inputenc} % allow utf-8 input
\usepackage[T1]{fontenc}    % use 8-bit T1 fonts
\usepackage{url}            % simple URL typesetting
\usepackage{booktabs}       % professional-quality tables
\usepackage{amsfonts}       % blackboard math symbols
\usepackage{nicefrac}       % compact symbols for 1/2, etc.
\usepackage{microtype}      % microtypography
\usepackage{xcolor}         % colors

\usepackage{float, soul, graphicx, bm, amsmath, amssymb, dsfont, multirow, amsthm}
\usepackage{amsmath}
\usepackage{amsfonts}
\usepackage{amssymb}
\usepackage[linesnumbered]{algorithm2e}
\usepackage{mathtools}
\RestyleAlgo{ruled}
\SetKwComment{Comment}{\# }{}
\newcommand{\R}{{\mathbb{R}}}
\newcommand{\Em}{{\mathbb{E}}}

\newcommand{\bx}{{\bm{x}}}

\newcommand{\p}{\mathsf{p}}
\newcommand{\n}{{\mathsf{n}}}

\newcommand{\gvn}{\,|\,}
\newcommand{\ds}{\displaystyle}

\renewcommand{\hat}{\widehat}
\newcommand{\I}{\mathds{1}}

\newcommand{\argmin}{\mathrm{argmin}}

\newtheorem{proposition}{Proposition}
\newcommand{\defeq}{{\coloneqq}}

\usepackage{wrapfig}
\usepackage{xspace}
\usepackage{thmtools, thm-restate}

\newcommand{\uPU}{{\rm uPU}\xspace}
\newcommand{\nnPU}{{\rm nnPU}\xspace}
\newcommand{\IR}{{\rm IR}}
\newcommand{\RR}{{\rm RR}}
\newcommand{\Gini}{{\rm gini}}
\newcommand{\Entropy}{{\rm entropy}}
\newcommand{\Quad}{{\rm quad}}
\newcommand{\Logistic}{{\rm logistic}}

\title{Positive-Unlabeled Learning using Random Forests via Recursive Greedy Risk Minimization}

\author{%
  Jonathan Wilton{$^1$}, Abigail M.~Y.~Koay{$^2$}, Ryan K.~L.~Ko{$^2$}, Miao Xu{$^{2,3}$}, Nan Ye{$^1$}
  \\
{$^1$}School of Mathematics and Physics, The University of Queensland\\
  {$^2$}School of Information Technology and Electrical Engineering, The University of Queensland\\
    {$^3$}RIKEN, Japan 103-0027 \\
}

\usepackage[bookmarks=false]{hyperref}       % hyperlinks
\hypersetup{
    colorlinks=false,
    pdfborder={0 0 0},
    pdfborderstyle={/S/U/W 0}
}
\usepackage{cleveref}

\begin{document}

\maketitle

\begin{abstract}
	The need to learn from positive and unlabeled data, or PU learning, arises in many
	applications and has attracted increasing interest.
	While random forests are known to perform well on many tasks with positive and
	negative data, recent PU algorithms are generally based on deep neural
	networks, and the potential of tree-based PU learning is under-explored.
	In this paper, we propose new random forest algorithms for PU-learning.
	Key to our approach is a new interpretation of decision tree algorithms for
	positive and negative data as \emph{recursive greedy risk minimization algorithms}.
	We extend this perspective to the PU setting to develop new decision tree
	learning algorithms that directly minimizes PU-data based estimators for the
	expected risk.
	This allows us to develop an efficient PU random forest algorithm, PU extra trees.
	Our approach features three desirable properties:
	it is robust to the choice of the loss function in the sense that various loss
	functions lead to the same decision trees;
	it requires little hyperparameter tuning as compared to neural network based
	PU learning;
	it supports a feature importance that directly measures a feature's
	contribution to risk minimization.
	Our algorithms demonstrate strong performance on several datasets.
	Our code is available at \url{https://github.com/puetpaper/PUExtraTrees}.
\end{abstract}
\raggedbottom

\section{Introduction}
Positive and unlabeled learning (PU learning) has attracted increasing interest
recently, due to its broad applications, including 
disease gene identification \cite{mordelet2011prodige}, 
landslide susceptibility modeling \cite{wu2021landslide}, 
inference of transcriptional regulatory network \cite{mordelet2014bagging}, cybersecurity \cite{wu2018hpsd}
and many others \cite{bekker2020learning}.

Existing algorithms generally reduce PU learning to solving one or more
supervised learning problems.
A naive approach is to simply learn a classifier by treating all unlabeled
examples as negative, but unlabeled examples can be either positive or
negative.
Various approaches have been developed to better handle the uncertainty on the
labels of the unlabeled examples.
A common approach follows an iterative two-step learning process: 
the first step identifies reliable negative examples, then the second step
learns a model using the positive examples, the reliable negative examples and
possibly unlabeled examples \cite{liu2003building,he2018instance}.
Another common approach transforms the PU dataset into a weighted fully labeled
dataset \cite{lee2003learning,elkan2008learning,du2014analysis,kiryo2017positive}.
The transformed dataset is often constructed so that the empirical risk on the
transformed dataset provides a good estimate of the true risk.
Depending on the assumptions on how PU data is generated, different
transformations have been developed.

While random forests are known to perform well for many supervised learning
problems and in principle tree-based learning methods can be applied in the
above approaches, little work has been done to explore the potential of
tree-based methods for PU learning.
On the other hand, current state-of-the-art PU learning methods are often based
on deep neural networks \cite{kiryo2017positive,chen2020self}.
This is partly due to the recent advances of deep learning, which allows
training of powerful neural network models with relatively little effort.
In addition, while it is easy to train a neural network to minimize different
loss functions, it is not clear how this can be done for tree-based methods
(without overfitting the dataset), because existing tree-based methods have not
been designed to directly minimize the loss.
In particular, it is not clear how this can be done for nonstandard risk
estimators such as the non-negative PU risk (\nnPU), which has already been
observed to work well with neural networks  \cite{kiryo2017positive}.

We fill this gap by providing a theoretically justified approach, which we call
\emph{recursive greedy risk minimization}, to learn random forests from PU data in this
paper.
Our contributions are as follows
\begin{itemize}
	\item A new interpretation to existing decision tree learning algorithm that
		links the impurity reduction based learning algorithms to empirical risk
		minimization.
	\item Efficient new decision tree and random forest algorithms that directly
		minimize PU-data based estimators for the expected risk.
	\item Our approach has three desirable properties:
		it is robust to the choice of the loss function in the sense that various loss
		functions lead to the same decision trees;
		it requires little hyperparameter tuning as compared to neural network based
		PU learning;
		it supports a feature importance that directly measures a feature's
		contribution to risk minimization.
	\item Our algorithms demonstrate strong performance on several datasets.
\end{itemize}

We briefly review decision tree learning and PU learning in
\Cref{sec:background},
introduce our recursive greedy risk minimization approach for decision tree
learning in \Cref{sec:recursive} and the PU version of extra trees in
\Cref{sec:puet},
present experiment results in \Cref{sec:experiments},
discuss related works in \Cref{sec:related}, and conclude in
\Cref{sec:conclusions}.

\section{Background} 
\label{sec:background}

\paragraph{Decision tree learning from positive and negative data}
A decision tree is generally constructed by recursively partitioning the
training set so that we can quickly obtain subsets which are more or less in the
same class.

\begin{wrapfigure}[12]{R}{0.55\textwidth}
    \hspace{0.2cm}
	\begin{minipage}{0.52\textwidth}
		\vspace{-0.8cm}
		\begin{algorithm}[H]
			Notations: $\kappa$ - node; $S$ - dataset; $f$ - feature; $t$ - threshold;\\
			\eIf{termination\ criterion\ is\ met }{
				Compute the prediction value at $\kappa$ using $S$;
			}{
				Choose an optimal split $(f, t)$; \\
				Create two child nodes $\kappa_{f > t}$ and $\kappa_{f \le t}$ for $\kappa$;\\
				$\texttt{LearnDT}(\kappa_{f > t}, S_{f > t})$; \\
				$\texttt{LearnDT}(\kappa_{f \le t}, S_{f \le t})$;
			}
			\caption{\texttt{LearnDT}($\kappa, S$)}
			\label{alg:learndt}
		\end{algorithm}
	\end{minipage}
\end{wrapfigure}
\Cref{alg:learndt} shows a generic decision tree learning algorithm.
We start with a single node associated with the entire set of training
examples.
When we encounter a node $\kappa$ associated with a set $S$ of training
examples, we compute the prediction value at $\kappa$ if the termination
criterion is met.
If not, we first compute an optimal split $(f, t)$ for the dataset $S$,
where a split $(f, t)$ tests whether a feature $f$ is larger than
a value $t$, and the quality of a split is usually measured by its impurity
reduction.
Based on the test's outcome, we create two child nodes $\kappa_{f > t}$ and
$\kappa_{f \le t}$, split $S$ into two subsets $S_{f > t}$ and $S_{f \le t}$,
and finally continue the learning process with $(\kappa_{f > t}, S_{f > t})$ and 
$(\kappa_{f \le t}, S_{f \le t})$.

Given an impurity measure $\text{Impurity}(S)$, a split $(f, t)$'s impurity
reduction is 
\begin{align}
	\IR(f, t; S) 
	\defeq 
	\text{Impurity}(S) 
	- 
	\frac{|S_{f > t}|}{|S|} \text{Impurity}(S_{f > t})
	-
	\frac{|S_{f \le t}|}{|S|} \text{Impurity}(S_{f \le t}).
\end{align}
Typically, the Gini impurity $G(S)$ and the entropy $H(S)$ are used as the
impurity measure:
\begin{align}
	G(S)
	&\defeq
	1 - q_{+}^{2} - q_{-}^{2}
	= 2 q_{+}(1 - q_{+}), 
	&
	H(S)
	&\defeq
	- q_{+} \ln q_{+} - q_{-} \ln q_{-},
\end{align}
where $q_{+}$ and $q_{-}$ are the proportions of positive and negative examples
in $S$ respectively.

We will use $\IR_{\text{Gini}}(f, t; S)$ and 
$\IR_{\text{entropy}}(f, t; S)$ to denote the impurity reduction when using the
Gini impurity and the entropy impurity respectively.

\paragraph{PU learning}
We assume that the input $\bx\in\R^d$ and label $y\in\{-1,+1\}$ follow an
unknown joint distribution $p(\bx,y) = p(\bx\gvn y)\,p(y)$.
In the fully supervised case, the training data is generally assumed to be
independently sampled from $p(\bx, y)$.
In the PU setting, one common assumption on the data generation mechanism
\cite{du2014analysis,du2015convex,kiryo2017positive} is that 
the positive examples $P = \{\bx_i^{(\p)}\}_{i=1}^{n_\p}$ are sampled
independently from the P marginal 
$p_\p(\bx) \defeq p(\bx \gvn y=1)$,
and the unlabeled examples 
% $U = \{\bx_i^{(\u)}\}_{i=1}^{n_\u}$ 
$U = \{{\bx}_{i}^{(\mathsf{u})}\}_{i=1}^{n_{\mathsf{u}}}$ 
are sampled independently from the marginal 
$p(\bx) = p(\bx, +1) + p(\bx, -1)$.
Clearly, we have 
$p(\bx) = \pi\, p_\p(\bx)+(1-\pi)\,p_\n(\bx)$,
where $\pi \defeq p(y=+1)$ is the positive rate and 
$p_\n(\bx) \defeq p(\bx\gvn y=-1)$ is the N marginal.

The objective is to learn a score function $g:\R^d \to \R$ (which corresponds to a
classifier that outputs $+1$ iff $g(\bx) > 0$) such that it minimizes its
expected risk
\begin{align}
    R(g) 
		&\defeq 
		\Em_{(\bx,y)\sim p(\bx,y)}\,\ell(g(\bx),y),
\end{align}
where $\ell: \R \times \{-1,+1\}\to\R$ is a loss function that gives the loss
$\ell(v, y)$ incurred by predicting a score $v$ when the true
label is $y$. 
We focus on the quadratic loss $\ell_{\text{quad}}(v, y) = (1-v y)^2$
, the logistic loss 
$\ell_{\text{logistic}}(v, y) = \ln(1+\exp(-v y))$
, the savage loss 
$\ell_{\text{savage}}(v, y) = 4/(1 + \exp({v y}))^{2}$
and the sigmoid loss
$\ell_{\text{sigmoid}}(v, y) = 1/(1+\exp(v y))$
in this paper. 

The risk can be written in terms of positive and unlabeled data as \cite{du2014analysis}
\begin{align}
	R(g) 
	&= 
	\pi\,\Em_{\bx\sim p_\p(\bx)}\,\ell(g(\bx),+1) + \Em_{\bx\sim p(\bx)}\,\ell(g(\bx),-1) - \pi\, \Em_{\bx\sim p_\p(\bx)}\,\ell(g(\bx),-1).
\end{align}
This gives the following unbiased PU-data based risk estimator
\begin{align}
    \hat{R}_{\uPU}(g) 
		&\defeq
		\sum_{\bx \in P} w_{\mathsf{p}} \ell(g(\bx), +1)
		-
		\sum_{\bx \in P} w_{\mathsf{p}} \ell(g(\bx), -1)
		+
		\sum_{\bx \in U} w_{\mathsf{u}} \ell(g(\bx), -1),
\end{align}
% where $w_{\p} = \pi/n_{\p}$ and $w_{\u} = 1/n_{\u}$.
where $w_{\mathsf{p}} = \pi/n_{\mathsf{p}}$ and $w_{\mathsf{u}} = 1/n_{\mathsf{u}}$.

While the \uPU risk estimator is unbiased, it has a negative component that provides extra incentive for the classifier to try hard to fit to the positive examples, thus potentially making the risk negative and leading to overfitting.
% it can lead to negative estimates and cause overfitting.
The non-negative risk estimator
\begin{align}
   \hat{R}_{\nnPU}(g) 
		&\defeq
		\sum_{\bx \in P} w_{\p} \ell(g(\bx), +1)
		+
		\max\left\{
			0,
			\sum_{\bx \in U} w_{\mathsf{u}} \ell(g(\bx), -1)
			-
			\sum_{\bx \in P} w_{\p} \ell(g(\bx), -1)
		\right\}
\end{align}
alleviates this problem by forcing the sum of the negative term and the term defined on the unlabeled data to be non-negative, as the sum acts as an estimate for risk on negative examples

\section{Recursive Greedy Risk Minimization} \label{sec:recursive}
\subsection{Decision trees for positive and negative data} 
We first introduce our recursive greedy risk minimization approach in the fully 
labeled case.
Consider the empirical risk of $g: \R^{d} \to \R$ on a labeled training set 
$D = \{(\bx_{1}, y_{1}), \ldots, (\bx_{n}, y_{n})\}$:
$\hat{R}(g) = \sum_{(\bx, y) \in D} w \ell(g(\bx), y)$, where 
$w = 1/|D|$.
If $g$ predicts a constant $v$ on a subset $S = \{(\bx, y)\}$ of the training
examples, then the contribution to the total empirical risk is the 
\emph{partial empirical risk}
$\hat{R}(v; S)
\defeq
\sum_{(\bx, y) \in S} w \ell(v, y)$.
The optimal constant value prediction is 
$v^{*}_{S} = \argmin_{v \in \R} \hat{R}^{*}(v; S)$
with a minimum partial empirical risk 
$\hat{R}^{*}(S) 
\defeq
\hat{R}^{*}(v^{*}_{S}; S)$.

If we switch from a constant value prediction rule to a decision stump
that uses the split $(f, t)$, then the minimum partial empirical risk for such a
decision stump is $\hat{R}^{*}(S_{f > t}) + \hat{R}^{*}(S_{f \le t})$, thus the 
\emph{risk reduction} for the split $(f, t)$ is
\begin{align}
	\RR(f, t; S) 
	\defeq 
	\hat{R}^{*}(S) 
	- 
	\hat{R}^{*}(S_{f > t})
	-
	\hat{R}^{*}(S_{f \le t}).
\end{align}

In the following, when we need to make the loss function explicit, we will use
subscript to indicate that.
For example, $\hat{R}^{*}_{\Quad}(S)$ and $\RR_{\Quad}(f, t; S)$, 
are the minimum partial empirical risk for a constant-valued prediction on $S$
and the risk reduction for the split $(f, t)$  on $S$ when using the quadratic loss.

When we choose the $(f, t)$ with maximum risk reduction in \Cref{alg:learndt},
we obtain a recursive greedy risk minimization algorithm that recursively zooms
in to an input region with a constant prediction, then improves the prediction
rule in a greedy manner by replacing it with the decision stump with minimal
empirical risk in that region.

The first main result shows that the Gini impurity reduction and entropy
reduction of a split are just scaled versions of the risk reductions when using
the quadratic loss and the logistic loss respectively.
This implies that the standard impurity reduction based decision learning
algorithm is in fact performing recursive greedy risk minimization.
All proofs are in the appendix.

\begin{restatable*}[]{thm}{pnir}
\label{thm:pnir}
	(a) For any $S \subseteq D$, we have 
			$\hat{R}^{*}_{\Quad}(S) = 2 |S| w G(S)$.
			As a consequence, for any $S \subseteq D$ and any split $(f, t)$ {\normalfont \cite{CART}},
			\begin{align}
				\RR_{\Quad}(f, t; S) = 2|S| w \IR_{\Gini}(f, t; S).
			\end{align}
	\noindent(b) For any $S \subseteq D$, we have 
			$\hat{R}^{*}_{\Logistic}(S) =  {|S|} w H(S)$.
			As a consequence, for any $S \subseteq D$ and any split $(f, t)$,
			\begin{align}
				\RR_{\Logistic}(f, t; S) = |S| w \IR_{\Entropy}(f, t; S).
			\end{align}
\end{restatable*}

\subsection{Decision trees for positive and unlabeled data}

We are now ready to introduce a recursive greedy risk minimization approach to
PU learning.
We do this by making two changes to \Cref{alg:learndt}:
the fully labeled dataset $S$ is replaced by a set $P' \subseteq P$ of positive
examples and a set $U' \subseteq U$ of unlabeled examples, and the split 
$(f, t)$ is chosen to optimize a PU version of the risk reduction.

\paragraph{\uPU risk reduction}
We first consider the simpler case of the \uPU estimator.
Similarly to the fully labeled case, if $g$ predicts a constant $v$ on $P'$ and
$U'$, then the contribution to the total empirical \uPU risk is
\begin{align*}
	\hat{R}_{\uPU}(v; P', U')
	\defeq
	\sum_{\bx \in P'} w_{\p} \ell(v, +1) 
	-
	\sum_{\bx \in P'} w_{\p} \ell(v, -1) 
	+
	\sum_{\bx \in U'} w_{\mathsf{u}} \ell(v, -1).
\end{align*}
The optimal constant value prediction is 
$v^{*}_{P', U'} = \argmin_{v \in \R} \hat{R}^{*}_{\uPU}(v; P', U')$
with a minimum empirical risk 
$\hat{R}^{*}_{\uPU}(P', U') 
\defeq
\hat{R}^{*}_{\uPU}(v^{*}_{P', U'}; P', U')$.
The \emph{\uPU risk reduction} for the split $(f, t)$ on $(P', U')$ is
\begin{align}
	\RR_{\uPU}(f, t; P', U') 
	\defeq 
	\hat{R}^{*}_{\uPU}(P', U') 
	- 
	\hat{R}^{*}_{\uPU}(P'_{f > t}, U'_{f > t})
	-
	\hat{R}^{*}_{\uPU}(P'_{f \le t}, U'_{f \le t}). \label{eq:RR}
\end{align}

The following result shows that the optimal risk $\hat{R}^{*}_{\uPU}(P', U')$
has a closed-form formula and is thus efficiently computable.
As a consequence, the \uPU risk reduction for a split $(f, t)$ is efficiently
computable.

\begin{restatable*}[]{prop}{uPUOptRisk}
\label{prop:uPUOptRisk}
	Consider arbitrary $P' \subseteq P$ and $U' \subseteq U'$.
	Let $W_{\p} = |P'| w_{\p}$, $W_{\n} = |U'| w_{\mathsf{u}} - |P'| w_{\p}$, and
	$v^{*} \defeq \frac{W_{\p}}{W_{\p} + W_{\n}}$.

	(a) When using the quadratic loss, we have 
	\begin{align}
		\hat{R}^{*}_{\uPU}(P', U')
		=
		\begin{cases}
			-\infty, &  v^{*} = +\infty, \\
			\ds 4 (W_{\p} + W_{\n}) v^{*} \left( 1-v^{*} \right), & \mathrm{otherwise}.
		\end{cases}
		\label{eq:gini1}
	\end{align}

	(b) When using the logistic loss, we have 
	\begin{align}
		\hat{R}^{*}_{\uPU}(P', U')
		=
		\begin{cases}
			(W_{\p} + W_{\n})\left(-v^{*}\ln(v^{*}) - (1-v^{*})\ln(1-v^{*})\right),& 0 < v^{*} < 1, \\
			0, & v^{*} \in \{0,1\}, \\
			-\infty, & v^{*} > 1.
		\end{cases}
		\label{eq:entropy1}
	\end{align}
\end{restatable*}

Intuitively, if $P' \subseteq P$ and $U' \subseteq U$ are the examples in a node
$\kappa$ in the decision tree, then the weights $W_{\p}$ and $W_{\n}$ serve as unbiased estimates of the probabilities 
$p(\bx \in \kappa, y = 1)$
and 
$p(\bx \in \kappa, y = -1)$ 
respectively. See \Cref{app:unbiased} for details.
Thus $v^{*} = \frac{W_{\p}}{W_{\p} + W_{\n}}$ serves as an estimate of the
probability of positive examples.
Note that $W_\p\in[0,1]$, while $W_\n\leq 1$ is unusual in that it contains a negative component and hence can be negative. 
Consequently, $v^{*}$ is non-negative, but it can be larger than 1 though.
In particular, if $W_{\p} + W_{\n} = |U'| w_{\mathsf{u}} = 0$, then $v^{*} = +\infty$.

\paragraph{\nnPU risk reduction}
For the case of the \nnPU estimator, define 
\begin{align*}
	\hat{R}_{\nnPU}(v; P', U')
	\defeq
	\sum_{\bx \in P'} w_{\p} \ell(v, +1) 
	+
	\max\left\{0,
	\sum_{\bx \in U'} w_{\mathsf{u}} \ell(v, -1)
		-
	\sum_{\bx \in P'} w_{\p} \ell(v, -1)\right\}.
\end{align*}
We can then similarly define the risk reduction for the \nnPU estimator.

The following result shows that in the \nnPU case, the optimal risk
$\hat{R}^{*}_{\uPU}(P', U')$ also has a closed-form formula and is thus
efficiently computable.

\begin{restatable*}[]{prop}{nnPUOptRisk}
\label{prop:nnPUOptRisk}
	Using the same notations as in 
	\Cref{prop:uPUOptRisk}, we have the following
	results.
	
	(a) When using the quadratic loss, we have 
	\begin{align}
		\hat{R}^{*}_{\nnPU}(P', U')
		=
		\begin{cases}
			0, &  v^{*} > 1\\
			\ds 4 (W_{\p} + W_{\n}) v^{*} \left( 1-v^{*} \right), & \mathrm{otherwise}.
		\end{cases}
	\end{align}

	(b) When using the logistic loss, we have 
	\begin{align}
		\hat{R}^{*}_{\nnPU}(P', U')
		=
		\begin{cases}
			(W_{\p} + W_{\n})\left(-v^{*}\ln(v^{*}) - (1-v^{*})\ln(1-v^{*})\right),& 0 < v^{*} < 1 \\
			0, & \mathrm{otherwise}.
		\end{cases}
	\end{align}
\end{restatable*}

It is important to note that while in the \uPU case, 
if $P'$ is partitioned into $P'_{1}$ and $P'_{2}$, and $U'$ into $U'_{1}$ and
$U'_{2}$, then we have 
$\hat{R}_{\uPU}(u; P', U') 
= 
\hat{R}_{\uPU}(u, P'_{1}, U'_{1})
+
\hat{R}_{\uPU}(u, P'_{2}, U'_{2})$, we only have 
$\hat{R}_{\nnPU}(u; P', U') 
\le 
\hat{R}_{\nnPU}(u, P'_{1}, U'_{1})
+
\hat{R}_{\nnPU}(u, P'_{2}, U'_{2})$.
Thus recursive greedy risk minimization minimizes an upper bound of the partial
empirical risk rather than the partial empirical risk.
This can be considered as a regularization mechanism.
In fact, we observe \nnPU risk to work better than the \uPU risk in our
experiments, which is consistent with the findings in \cite{kiryo2017positive}.

\paragraph{Optimizing the split}
With the above closed-form formula, we can then efficiently compute the risk
reduction for a given split $(f, t)$ on $(P', U')$. 
While there are infinitely many possible $t$ values to choose from, note that
they often split the dataset in the same way, and if $f$'s values split $\R$ into
multiple intervals, we only need to consider $t_{1} < t_{2} < \ldots < t_{m}$, each
chosen from an interval so that we cover all possible cases.
While computing the risk reduction for each $(f, t_{i})$ takes $O(|P'| + |U'|)$ time,
we can efficiently get the risk reductions for all of $(f, t_{1}), \ldots, (f, t_{m})$
in $O((|P'| + |U'|)\ln(|P'| + |U'|) + m)$ time, by sorting the examples
according to their $f$ values, and going through the thresholds in a sorted
order.
See \Cref{app:optimising_split} for details.

\paragraph{Invariance properties}
An interesting property of the greedy recursive learning approach is that it is
robust to the choice of the loss function in the sense that different loss
functions can lead to identical decision trees, as shown in the result below.
\begin{restatable*}[]{prop}{quadsavage}
\label{prop:quadsavage}
	The $\hat{R}^{*}(S)$ value is the same for the quadratic loss and
	the savage loss.
\end{restatable*}

Another interesting property is that the optimal prediction among $-1$ and $+1$
are often the same when we use different estimators and different loss
functions.

\begin{restatable*}[]{prop}{optimalpredictionfunction}
\label{prop:optimal_prediction_function}
	If a node contains the examples $P' \subseteq P$ and $U' \subseteq U$, 
	$v^{*}$ is as defined in \Cref{prop:uPUOptRisk}, and 
	the prediction is chosen from $\{-1, +1\}$ to minimize the empirical risk estimate
	on $P'$ and $U'$, then the optimal prediction is 
	$2\I(v^{*} > 0.5) - 1$ for both the \uPU and \nnPU risk
	estimate and all the loss functions in \Cref{sec:background}.
	Intuitively, we predict $+1$ iff the estimated proportion of positive data at the
	current node is larger than 0.5. 
\end{restatable*}

\section{PU Extra Trees} \label{sec:puet}

With our PU decision tree algorithms, we can obtain a PU random forest algorithm
by adapting the random forest algorithm \cite{breiman2001random} for positive
and negative data:
repeatedly obtain bootstrap samples for both the positive data and the unlabeled
data, then train a PU decision tree for each sample with optimal split chosen
over a randomly sampled subset of features each time.
While choosing the optimal split from a subset of features makes training each
decision tree faster, it is still the most expensive part in the tree
construction process and can potentially be very slow on a large dataset.
We describe a more efficient version that we used in our experiments below.
In addition, we introduce a new feature importance score that directly measures
a feature's contribution to risk minimization.

\paragraph{PU ET (Extra Trees)}
We develop a more efficient random forest algorithm by using the randomization
trick in the Extra Trees algorithm \cite{geurts2006extremely} to further reduce
the computational cost for finding an optimal split below.
Besides sampling only a subset of features, only one random threshold for each
sampled feature is considered.
We implemented a more general version which allows sampling multiple random
thresholds for a sampled feature.
A complete pseudocode for PU ET is given in \Cref{app:pseudocode}.

\paragraph{Termination criteria}
Decision trees can overfit the training data if allowed to grow until each leaf
has just one example.
Learning is generally terminated early to alleviate overfitting.
In our implementation, we terminate when all feature values are constant, 
when a maximum tree depth $d_{\max}$ is reached, 
when the node size falls below a threshold $s_{\min}$,
or when the node is \emph{pure}.
A node is said to be pure if the impurity measure takes value $-\infty$ in the \uPU setting, 
or value 0 in the \nnPU setting. 

\paragraph{Risk reduction importance} 
Our 
% new 
feature importance, called \emph{risk reduction importance}, is the total risk
reduction for a feature across all nodes in all the trees. 
The risk reduction importance of a feature $f$ is defined as $\sum_{\kappa\in K_f} \RR(f_\kappa,t_\kappa;P_\kappa,U_\kappa)$, where $K_f$ is the set of nodes using $f$ as a splitting feature, $(f_\kappa,t_\kappa)$ is the split for node $\kappa$, and $(P_\kappa,U_\kappa)$ is the set of PU data at $\kappa$. This can be averaged across multiple trees in a random forest.  
The risk reduction importance in PU learning is similar to the Gini importance as defined in, for example, \cite{CART}, whereby the size of the nodes is taken into account. This is beneficial as otherwise features often used in splitting small datasets may appear very important as compared to features used in splitting large datasets, even though it does not contribute as much to reducing the risk and improving classification performance. 
To demonstrate the effect of taking data size into account, a normalized risk reduction can be defined by dividing the risk reduction by the total weight of the examples. 

\section{Experiments}\label{sec:experiments}
In this section we compare PU Extra Trees with several other PU learning methods. Selected tree-based methods include NaivePUET whereby ET is trained on the PU dataset by simply treating U data as N data; PUBaggingET which uses PUBagging from \cite{mordelet2014bagging} with ET base classifier; and SupervisedET where ET is trained on the original fully labeled dataset (expected to be an upper bound for tree based PU methods). To further study the effectiveness of our PU Extra Trees algorithm we also compare against neural network based methods including a baseline method \uPU \cite{du2014analysis, du2015convex} and two state of the art methods \nnPU \cite{kiryo2017positive} and Self-PU \cite{chen2020self}. 
We ran experiments on heterogeneous devices. In particular, random forests were trained using 32GB RAM and one of Intel i7-10700, Intel i7-11700 or AMD Epyc 7702p CPU. Neural networks were trained on one of NVIDIA RTX A4000 or NVIDIA RTX A6000 GPU due to the lack of identical devices. While the running times are not always comparable and thus not provided in the main text, experiments show that our current implementation of PU ET takes seconds to train a single-tree random forest on modest hardware. See \Cref{app:times} for details.

\vspace{-.5em}
\paragraph{Datasets}
We consider a selection of common datasets for classification from LIBSVM \cite{libsvm}, as well as MNIST digits \cite{mnist}, the intrusion detection dataset UNSW-NB15 \cite{unsw1} and CIFAR-10 \cite{cifar-10} to demonstrate the versatility of our method. 
Table \ref{tab:datasets} is a summary of the benchmark datasets. 
\begin{table}[H]
    \centering
    \caption{Benchmark datasets. 
		$^*$: random 80\%-20\% train-test split was used as no train-test splits were provided.}
    \begin{tabular}{lcccc}
        Name & \# Train & \# Test & \# Feature & $\pi$ \\ \hline
        mushrooms$^*$ & 6499 & 1625 & 112 & 0.52 \\
        % a9a & 32 561 & 16 281 & 123 & 0.24 \\
        20News & 11 314 & 7 532 & 300 & 0.56 \\
        % w8a & 49 749 & 14 951 & 300 & 0.03 \\
        covtype.binary$^*$ & 464 809 & 116 203 & 54 & 0.51 \\
        epsilon & 400 000 & 100 000 & 2 000 & 0.5   \\
        % HIGGS & 10 500 000 & 500 000 & 28 & 0.53 \\
        MNIST & 60 000 & 10 000 & 784 & 0.5 \\
        CIFAR-10 & 50 000 & 10 000 & 3 072 & 0.4 \\
        UNSW-NB15 & 175 340 & 82 331 & 39 & 0.68
    \end{tabular}
    \label{tab:datasets}
		\vspace{-.5em}
\end{table}
The 20News, epsilon, MNIST and CIFAR-10 datasets were processed in the same way as in \cite{kiryo2017positive} for consistency. 
Definitions of labels (`positive' vs `negative') are as follows: For 20News, ‘alt., comp., misc., rec.’ vs ‘sci., soc., talk.’; for MNIST, `0,2,4,6,8' vs `1,3,5,7,9'; for CIFAR-10, ‘airplane, automobile, ship, truck’ vs ‘bird, cat, deer, dog, frog, horse’; for UNSW-NB15, all attack types make up the P class and the benign data makes up the N class; for mushroom and covtype, the most prominent class makes up the P class; and epsilon dataset is provided with P and N classes. 
GloVe pre-trained word embeddings \cite{glove} were used for 20News with average pooling over each document.
Each feature in the covtype and UNSW-NB15 datasets was scaled between 0 and 1 when using \uPU and \nnPU as there is a large mismatch in scales between the features which seemed to significantly reduce performance of the methods. No such scaling of the data is required for PU ET as performance does not seem to be affected in such cases. 

To convert the PN data to PU data we follow the experimental setup in \cite{kiryo2017positive} whereby 1000 positive examples are randomly sampled to form the P set, and the entire dataset is used for the U set. In practice it may be the
case that $\pi$ is not known a-priori, in which case there are methods for
estimating $\pi$ using only PU data, for example, in \cite{prior-estimation1}, or \cite{prior-estimation2}. 

\vspace{-.5em}
\paragraph{Model hyperparameters}
Following common practice \cite{sklearn,geurts2006extremely}, the default hyperparameters for PU ET are: 
100 trees, 
no explicit restriction on the maximum tree depth, 
sample $F = \lceil \sqrt{d} \rceil$ features out of a total of $d$ features and
sample $T=1$ threshold value when computing an optimal split.
The size of the bootstrap sample for PUBaggingET is set to the default value of the dataset size. 

The architectures for the neural networks used in \uPU, \nnPU and Self-PU were
copied from \cite{kiryo2017positive} for the 20News, epsilon, MNIST and CIFAR-10 datasets. A 6 layer MLP with ReLU was used for MNIST, Covtype, Mushroom and UNSW-NB15; a similar model was used for epsilon while the activation was replaced with Softsign; a 5 layer MLP with Softsign was used for 20News. All hidden layers had width 300 for the MLPs. The model for CIFAR-10 was the 13 layer CNN: (32*32*3,1)-[C(3*3,96,1)]*2-C(3*3,96,2)-[C(3*3,192,1)]*2-C(3*3,192,2)-C(3*3,192,1)-C(1*1,192,1)-C(1*1,10,1)-1000-1000-1, where the input is a 32*32 RGB image, C(3*3,96,1) means 96 channels of 3*3 convolutions with stride 1 followed by ReLU, [.]*2 means there are two such layers. For the remaining datasets we used a 6-layer MLP with ReLU \cite{relu1, relu2} (more specifically, $d$-300-300-300-300-1). 
For each dataset the neural networks were trained for 200 epochs. 
The batch size, learning rate, use of batch-norm (for datasets not included in
\cite{kiryo2017positive}), weight decay and choice of optimiser were tuned for
each dataset. 

It is noteworthy that we spend little effort to tune PU ET hyperparameters and simply use the same hyperparameters on all datasets. In addition, PU ET seems to be robust to hyperparameter choice in that the default selection tends to result in strong predictive performance with decreased training times. See \Cref{app:hyperparams} for results from hyperparameter tuning experiment. On the other hand, neural nets are sensitive to hyperparameter choice, and significant tuning is often needed for each dataset.

\subsection{Classification performance}
To take into account of the randomness in the learning algorithms, we train each model five times on the same dataset, then report the mean and standard deviation of the accuracy and the F-score.
Results for \uPU, \nnPU and Self-PU were reproduced to the best of our ability, particularly on the datasets that were included in the original results section of the respective papers. 

\paragraph{Effect of loss function and risk estimator on PU ET}
We first evaluate the effect of different loss functions and risk estimators for PU ET.
The results are shown in \Cref{tbl:acc_vs_risk} and \Cref{tbl:f_vs_risk}.
PU ET with nonnegative risk estimator seems to perform about as well using either quadratic or logistic loss.
Unbiased risk estimator seems to lead to overfitting, but less so with logistic loss.
Logistic loss leads to better performance than quadratic loss when using unbiased risk estimator. 
Note that \eqref{eq:gini1} implies that quadratic loss encourages aggressive splitting until all examples are positive as a loss of $-\infty$ can be achieved in this case, while \eqref{eq:entropy1} implies that logistic loss has a less aggressive splitting behavior as a loss of $-\infty$ is achieved when the total weight of positive examples exceeds the total weight of unlabeled examples. 
As discussed previously, once we achieve an impurity value of $-\infty$ in \uPU, we stop splitting. 
The more aggressive splitting behavior of quadratic loss leads to deeper trees that tend to overfit more than the logistic loss in general. This is consistent with results from additional experiments in \Cref{app:overfitting}.
\begin{table}[!ht]
    \centering
    \caption{Accuracy mean\% (sd) on the test set for PU ET using various PU data based risk estimators.}
    \begin{tabular}{lcccc}
        \multirow{2}{*}{Dataset} & \multicolumn{2}{c}{Unbiased risk estimator} & \multicolumn{2}{c}{Nonnegative risk estimator} \\  
          & Quadratic Loss & Logistic Loss & Quadratic Loss & Logistic Loss \\ \hline
         Epsilon   & $50.04\ (0.00)$ & $50.67\ (0.08)$ & $57.39\ (0.76)$ & $57.83\ (0.70)$ \\
         20News    & $43.64\ (0.05)$ & $73.44\ (0.87)$ & $83.34\ (0.22)$ & $83.34\ (0.31)$ \\
         Covtype   & $48.72\ (0.00)$ & $72.63\ (0.87)$ & $76.51\ (0.52)$ & $75.63\ (0.52)$ \\
         Mushroom  & $0.607\ (0.48)$ & $99.02\ (0.32)$ & $ 99.70\ (0.24)$ & $99.36\ (0.48)$ \\
				 MNIST     & $50.74\ (0.0)$ & $89.05\ (0.61)$ & $93.60\ (0.39)$ & $94.01\ (0.19)$ \\
         CIFAR-10  & $60.0.(0.01)$ & $75.30\ (0.66)$ & $79.74\ (0.37)$ & $79.86\ (0.39)$ \\
         UNSW-NB15 & $47.18\ (0.15)$ & $83.89\ (1.03)$ & $82.24\ (0.86)$ & $81.51\ (1.58)$ \\
    \end{tabular}
		\label{tbl:acc_vs_risk}
		\vspace{-1em}
\end{table}

\begin{table}[!ht]
    \centering
    \caption{F mean\% (sd) on the test set for PU ET using various PU data based risk estimators.}
    \begin{tabular}{lcccc}
        \multirow{2}{*}{Dataset} & \multicolumn{2}{c}{Unbiased risk estimator} & \multicolumn{2}{c}{Nonnegative risk estimator} \\  
          & Quadratic Loss & Logistic Loss & Quadratic Loss & Logistic Loss \\ \hline
         Epsilon   & $0.00\ (0.00)$ & $4.18\ (0.59)$ & $39.52\ (1.19)$ & $40.25\ (1.38)$ \\
         20News    & $0.43\ (0.17)$ & $72.65\ (1.11)$ & $85.33\ (0.22)$ & $85.26\ (0.28)$ \\
         Covtype   & $0.11\ (0.01)$ & $68.44\ (1.25)$ & $75.19\ (0.37)$ & $74.20\ (0.58)$ \\
         Mushroom  & $39.02\ (1.19)$ & $99.04\ (0.32)$ & $99.71\ (0.24)$ & $99.38\ (0.47)$ \\
				 MNIST     & $0.00\ (0.00)$ & $88.06\ (0.74)$ & $93.49\ (0.42)$ & $93.90\ (0.19)$ \\
         CIFAR-10  & $0.02\ (0.04)$ & $61.33\ (1.44)$ & $71.31\ (0.43)$ & $71.67\ (0.53)$ \\
         UNSW-NB15 & $7.97\ (0.47)$ & $86.33\ (0.87)$ & $85.65\ (0.59)$ & $85.17\ (1.15)$ \\
    \end{tabular}
		\label{tbl:f_vs_risk}
		\vspace{-1em}
\end{table}

\paragraph{Comparison of PU ET with other tree-based PU learning methods} 
We now compare our PU ET algorithm with several tree-based baselines to further study the effectiveness of our novel PU tree learning method. The results are shown in \Cref{tbl:acc_vs_tree} and \Cref{tbl:f_vs_tree}. 
The results support the effectiveness of our algorithm.
Our PU ET with \nnPU risk estimator and quadratic loss significantly outperforms both NaivePUET and PUBaggingET, particularly in terms of F-scores.
PU ET shows strong performance even as compared to SupervisedET, though often using a small proportion of the positive labels only.

\begin{table}[h!]
    \centering
    \caption{Accuracy mean\% (sd) on the test set for various tree based methods.
            $^\dagger$: Original dataset with full supervision was used during training.
            }
    \begin{tabular}{lcccc}
        Dataset & SupervisedET$^\dagger$    & PU ET        & NaivePUET    & PUBaggingET \\ \hline
        Epsilon   & 73.55 (0.08) & 57.39 (0.76) & 50.04 (0.00) & 50.04 (0.00)\\
        20news    & 85.39 (0.12) & 83.34 (0.22) & 43.63 (0.05) & 43.75 (0.06)\\
        Covtype   & 95.90 (0.02) & 76.51 (0.52) & 48.71 (0.00) & 48.71 (0.0)\\
        Mushroom  & 100 (0.00)   & 99.70 (0.24) & 53.85 (0.32) & 62.76 (0.84)\\
        MNIST     & 98.11 (0.05) & 93.60 (0.39) & 50.74 (0.00) & 50.74 (0.00)\\
        CIFAR-10  & 85.02 (0.19) & 79.74 (0.37) & 60.00 (0.00) & 60.0 (0.00)\\
        UNSW-NB15 & 86.57 (0.05) & 82.24 (0.86) & 44.95 (0.01) & 44.97 (0.02)
    \end{tabular}
		\label{tbl:acc_vs_tree}
\end{table}

\begin{table}[h!]
    \centering
    \caption{F mean\% (sd) on the test set for PU ET using various PU data based risk estimators.
                $^\dagger$: Original dataset with full supervision was used during training.
                }
    \begin{tabular}{lcccc}
        Dataset   & SupervisedET$^\dagger$  & PU ET            & NaivePUET  & PUBaggingET \\ \hline
        Epsilon   & 72.96 (0.08) & 39.52 (1.19) & 0.00 (0.00) & 0.00 (0.00) \\
        20news    & 87.40 (0.11) & 85.33 (0.22) & 0.38 (0.19) & 0.82 (0.18)\\
        Covtype   & 95.97 (0.02) & 75.19 (0.37) & 0.05 (0.01) & 0.06 (0.01)\\
        Mushroom  & 100 (0.00)   & 99.71 (0.24) & 19.68 (2.04) & 45.3 (0.84)\\
        MNIST     & 98.09 (0.05) & 93.49 (0.42) & 0.00 (0.00) & 0.00 (0.00)\\
        CIFAR-10  & 80.30 (0.24) & 71.31 (0.43) & 0.01 (0.02) & 0.02 (0.02)\\
        UNSW-NB15 & 88.96 (0.04) & 85.65 (0.59) & 0.08 (0.06) & 0.11 (0.06)
    \end{tabular}
		\label{tbl:f_vs_tree}
\end{table}

\paragraph{Comparison of PU ET with neural network methods}
We compare PU ET with \nnPU risk estimator and quadratic loss against \uPU neural nets, \nnPU neural nets and Self-PU. The results are shown in \Cref{tbl:acc} and \Cref{tbl:f}. PU ET shows strong performance in terms of both accuracy and F score on a wide variety of datasets compared to neural network based models. 

\begin{table}[!ht]
    \centering
    \caption{Accuracy\% mean (sd) on the test set for PU ET with nonnegative risk estimator and quadratic loss and three neural           network based PU learning methods.
            $^*$: Self-PU was run without self-calibration due to limitations with available hardware.
    }
    \begin{tabular}{lcccc}
        \multirow{2}{*}{Dataset} & \multirow{2}{*}{PU ET} & \multicolumn{3}{c}{Neural Network} \\ \cline{3-5}
                 & \multicolumn{1}{c}{} & \uPU & \nnPU & Self-PU \\ \hline
         Epsilon & $57.39\ (0.76)$ & $61.79\ (2.57)$ & $73.80\ (0.49)$ & $50.05\ (0.0)$ \\
         20News & $83.34\ (0.22)$ & $44.95\ (1.09)$ & $74.11\ (3.65)$ & $62.87\ (2.89)$ \\
         Covtype & $76.51\ (0.52)$ & $72.38\ (1.23)$ & $71.64\ (0.66)$ & $54.67\ (0.40)$ \\
         Mushroom & $99.70\ (0.24)$ & $98.09\ (1.03)$ & $99.04\ (0.54)$ & $99.25\ (0.61)$ \\
         MNIST & $93.60\ (0.39)$ & $55.95\ (1.17)$ & $93.08\ (0.46)$ & $93.13\ (0.41)$ \\
         CIFAR-10 & $79.74\ (0.37)$ & $62.68\ (2.95)$ & $81.87\ (4.20)$  & $88.22\ (0.51)^*$ \\
         UNSW-NB15 & $82.24\ (0.86)$ & $76.62\ (0.01)$ & $76.63\ (0.02)$ & $76.73\ (0.18)$
    \end{tabular}
    \label{tbl:acc}
\end{table}

\begin{table}[!ht]
    \centering
    \caption{F mean\% (sd) on the test set for PU ET with nonnegative risk estimator and quadratic loss and three neural network based PU learning methods.
		$^*$: Self-PU was run without self-calibration due to limitations with available hardware.}
    \begin{tabular}{lcccc}
        \multirow{2}{*}{Dataset} & \multirow{2}{*}{PU ET} & \multicolumn{3}{c}{Neural Network} \\ \cline{3-5}
                 & \multicolumn{1}{c}{} & \uPU & \nnPU & Self-PU \\ \hline
         Epsilon & $39.52\ (1.19)$ & $43.44\ (8.14)$ & $71.53\ (0.46)$ & $0.00\ (0.00)$ \\
         20News & $85.33\ (0.22)$ & $12.95\ (3.82)$ & $78.05\ (3.81)$ & $71.78\ (6.85)$ \\
         Covtype & $75.19\ (0.37)$ & $69.56\ (1.77)$ & $68.29\ (0.93)$ & $0.501\ (2.55)$ \\
         Mushroom & $99.71\ (0.24)$ & $98.13\ (1.02)$ & $99.06\ (0.53)$ & $99.27\ (0.60)$ \\
         MNIST & $93.49\ (0.42)$ & $19.13\ (3.82)$ & $92.77\ (0.60)$ & $92.96\ (0.47)$ \\
         CIFAR-10 & $71.31\ (0.43)$ & $32.43\ (25.03)$ & $76.01\ (7.62)$ & $85.51\ (0.60)^*$ \\
         UNSW-NB15 & $85.65\ (0.59)$ & $82.47\ (0.01)$ & $82.48\ (0.02)$ & $80.18\ (0.28)$
    \end{tabular}
    \label{tbl:f}
\end{table}

\subsection{Feature Importance}
\begin{figure}[H]
    \centering
    \begin{minipage}{0.49\textwidth}
        \centering
        \includegraphics[height = 5.8cm]{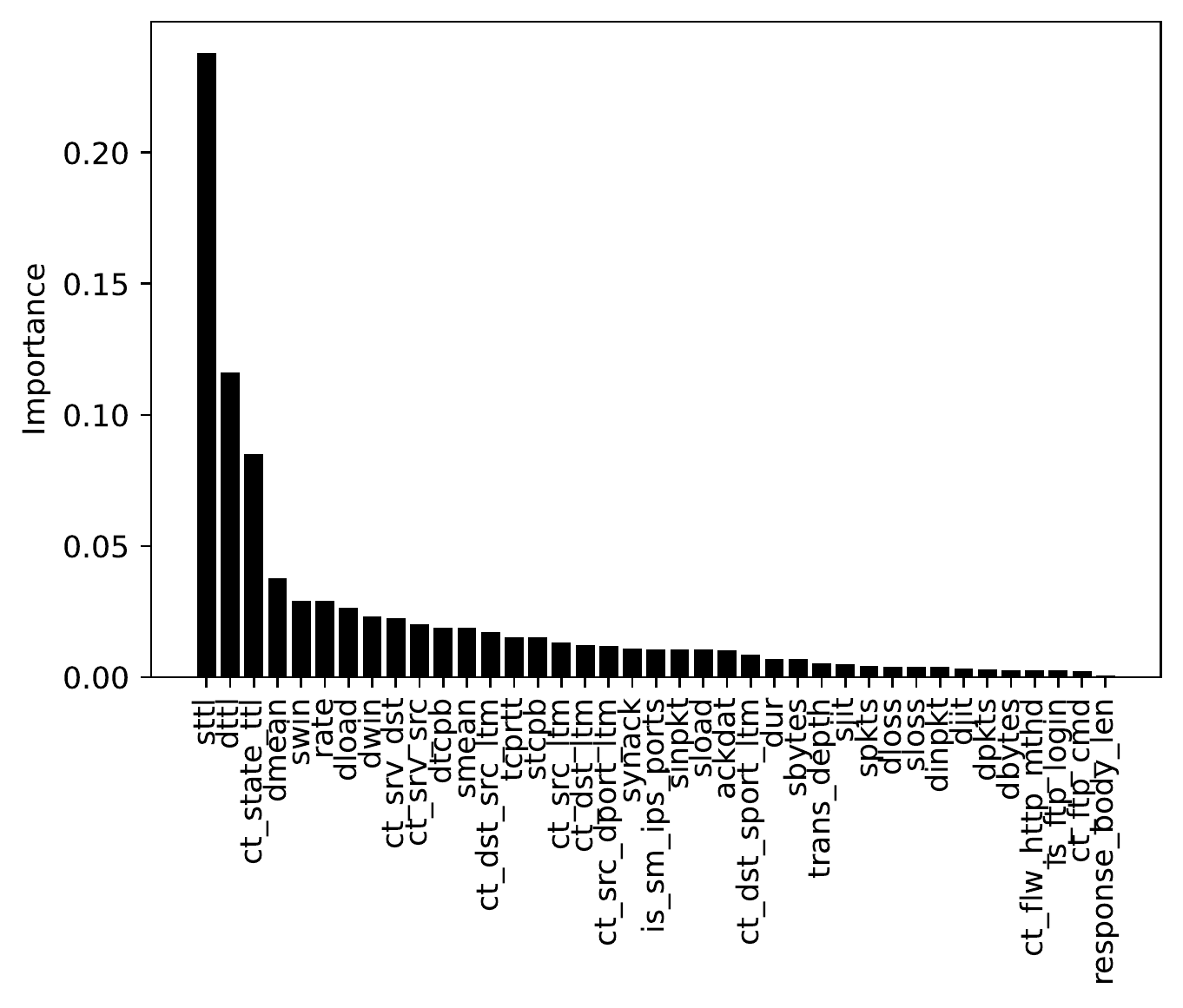}
    \end{minipage}
    \begin{minipage}{0.49\textwidth}
        \centering
        \includegraphics[height = 5.8cm]{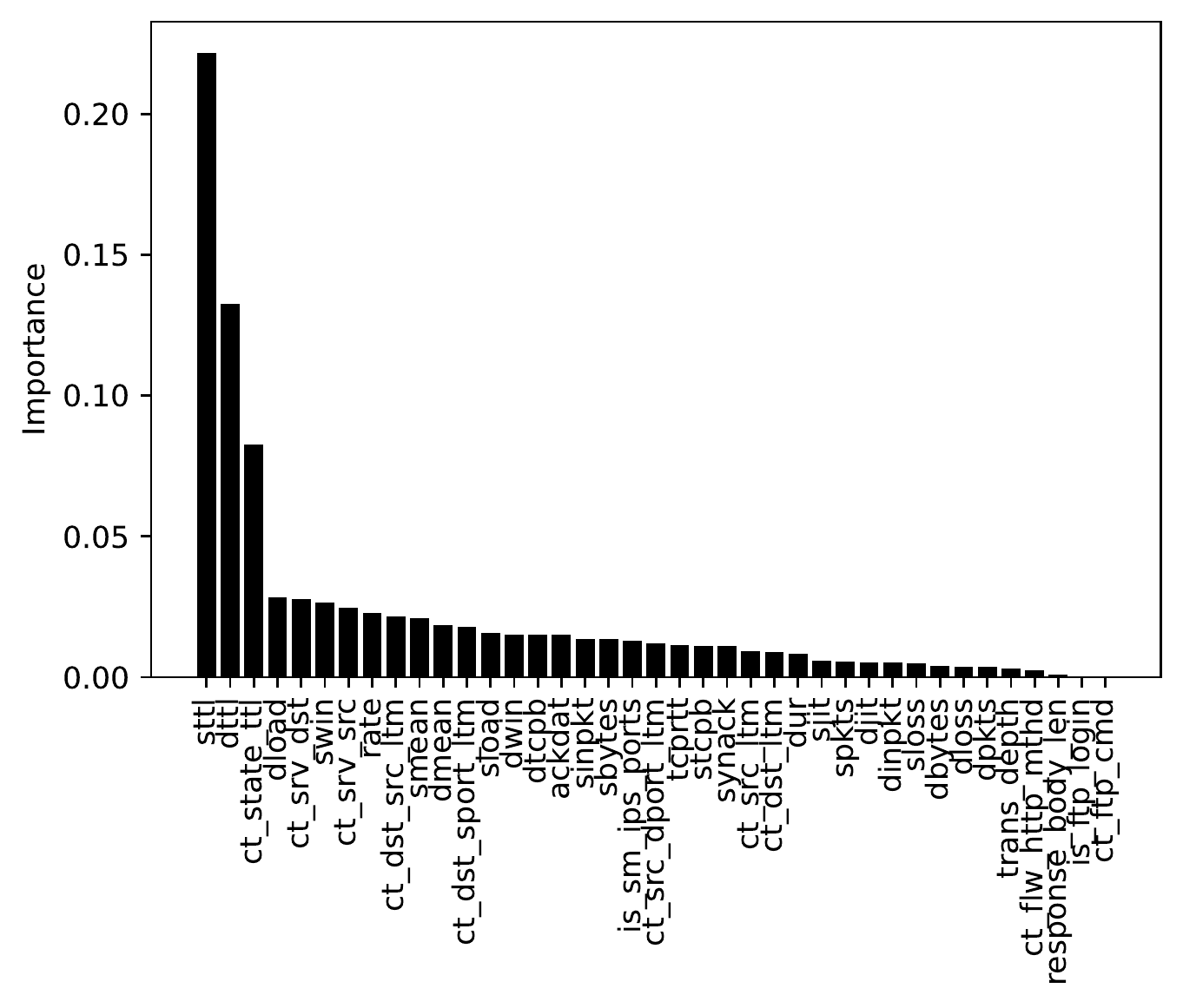}
    \end{minipage}
    \caption{Risk reduction importances on UNSW-NB15. Left: PU ET. Right: PN ET.}
    \label{fig:unsw_importances}
\end{figure}

\begin{figure}[h!]
    \centering
    \begin{minipage}{0.09\textwidth}
        \centering
        \includegraphics[scale = 0.15]{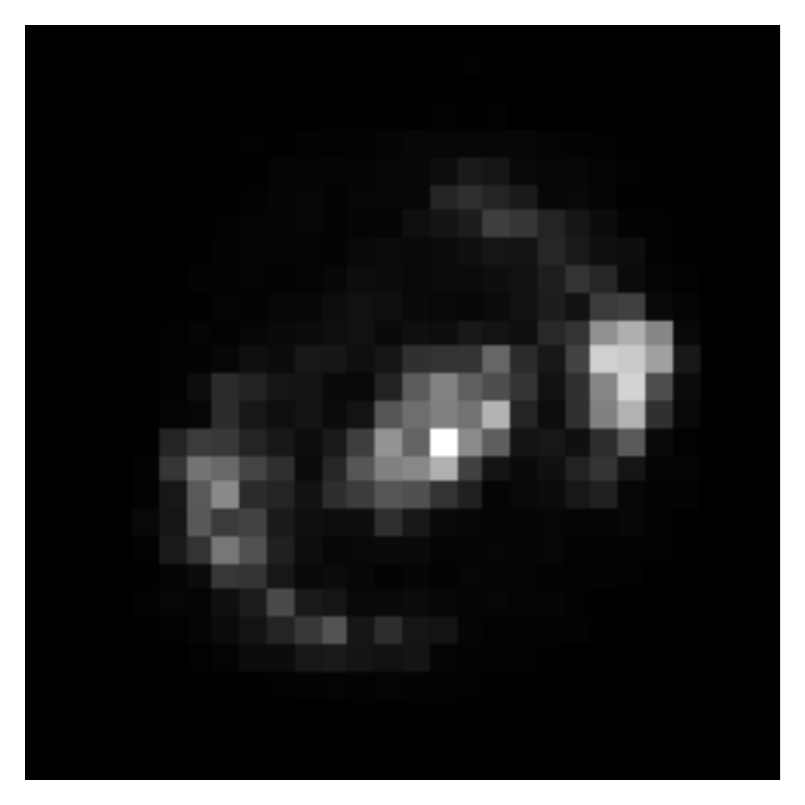}
    \end{minipage}
    \begin{minipage}{0.09\textwidth}
        \centering
        \includegraphics[scale = 0.15]{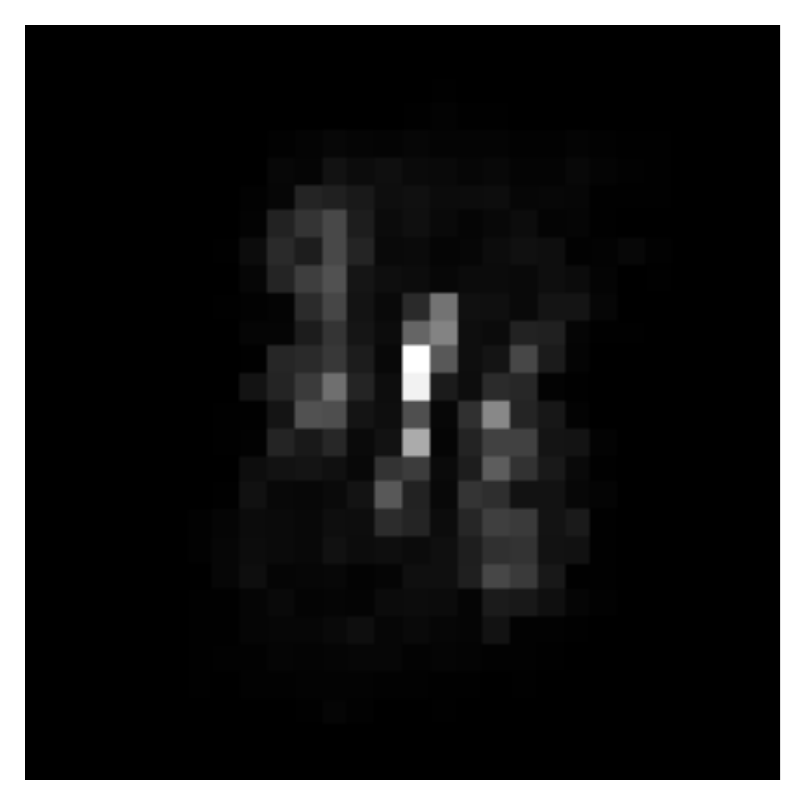}
    \end{minipage}
    \begin{minipage}{0.09\textwidth}
        \centering
        \includegraphics[scale = 0.15]{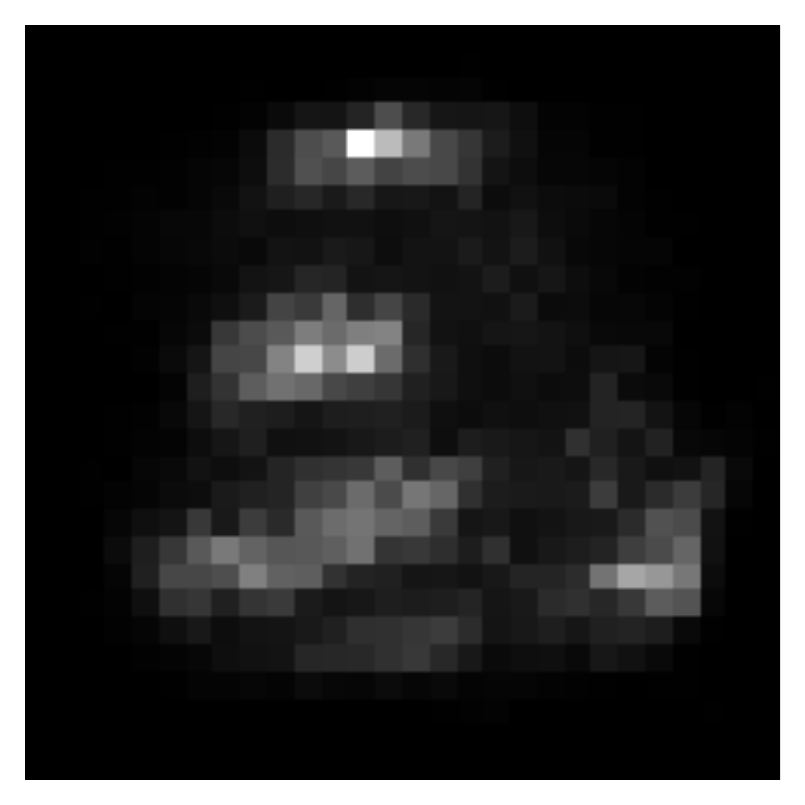}
    \end{minipage}
    \begin{minipage}{0.09\textwidth}
        \centering
        \includegraphics[scale = 0.15]{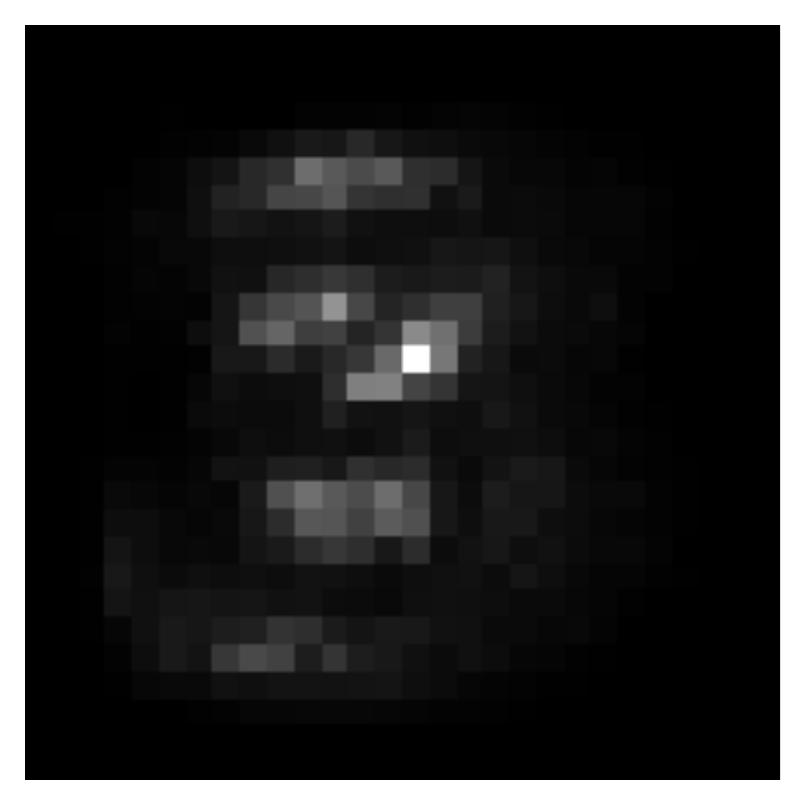}
    \end{minipage}
    \begin{minipage}{0.09\textwidth}
        \centering
        \includegraphics[scale = 0.15]{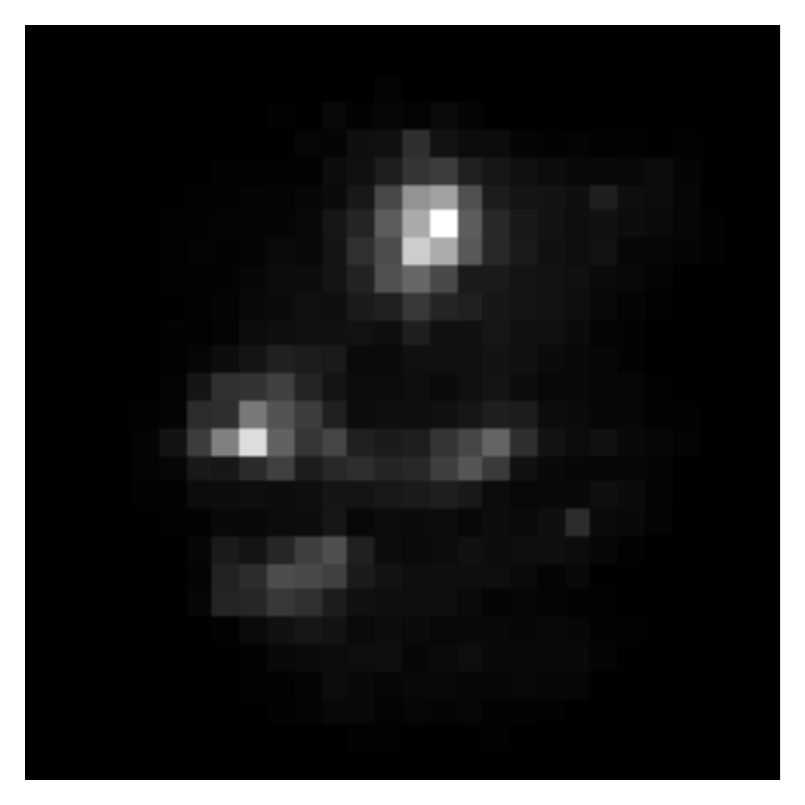}
    \end{minipage}
    \begin{minipage}{0.09\textwidth}
        \centering
        \includegraphics[scale = 0.15]{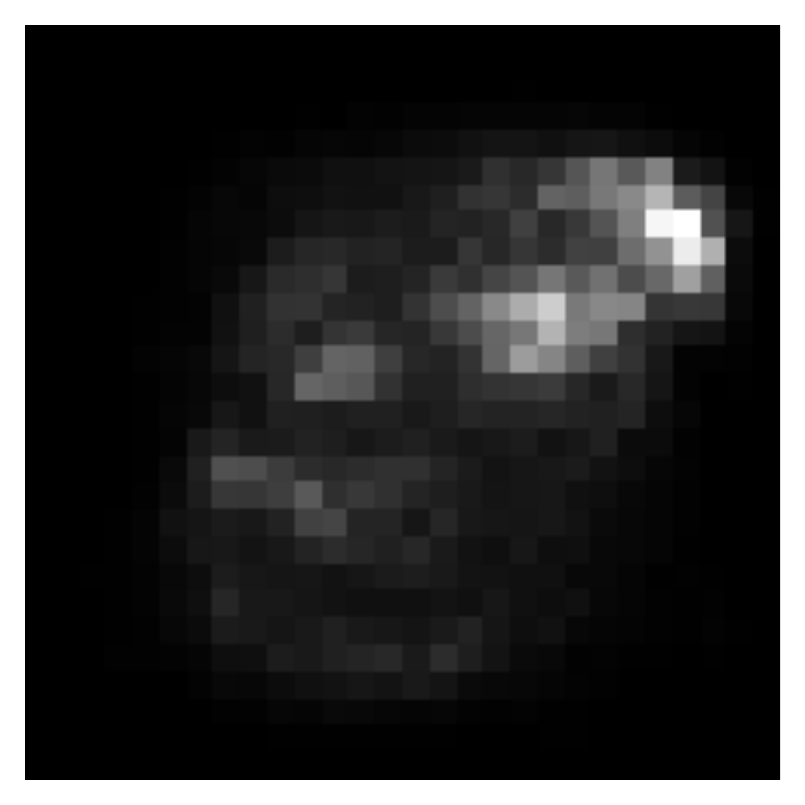}
    \end{minipage}
    \begin{minipage}{0.09\textwidth}
        \centering
        \includegraphics[scale = 0.15]{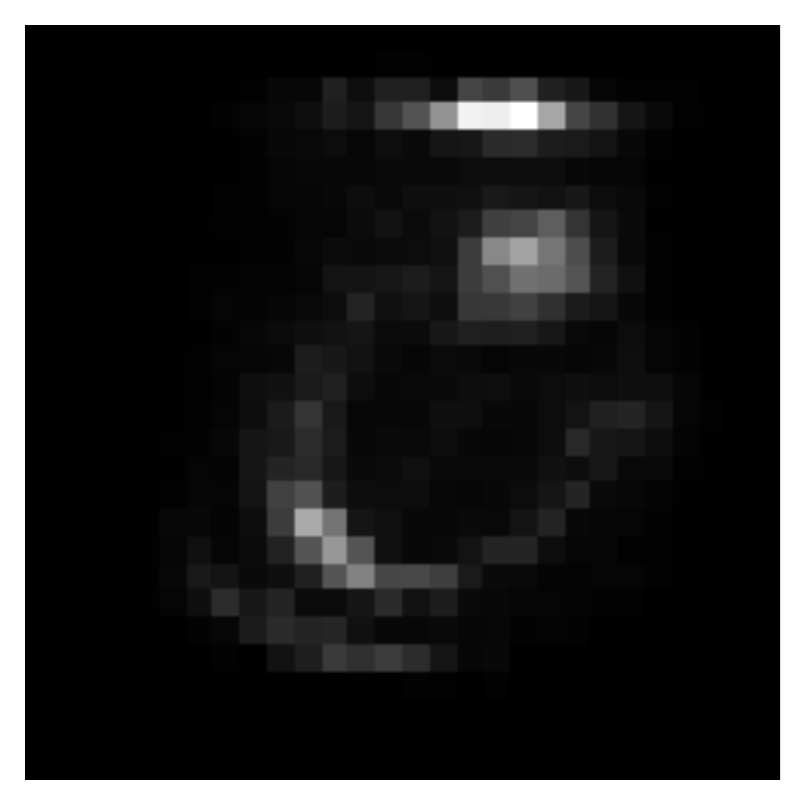}
    \end{minipage}
    \begin{minipage}{0.09\textwidth}
        \centering
        \includegraphics[scale = 0.15]{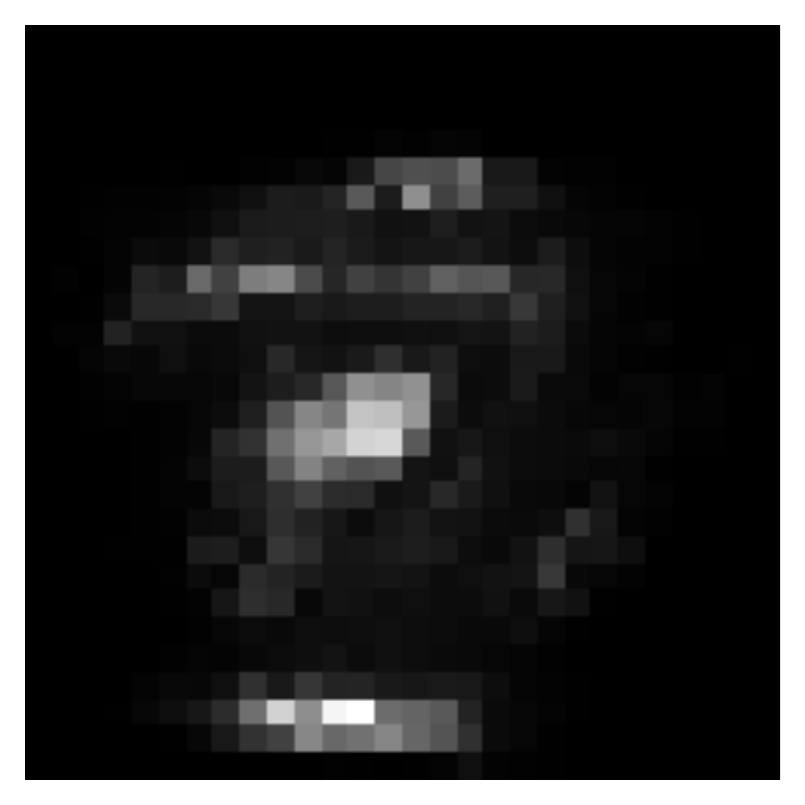}
    \end{minipage}
    \begin{minipage}{0.09\textwidth}
        \centering
        \includegraphics[scale = 0.15]{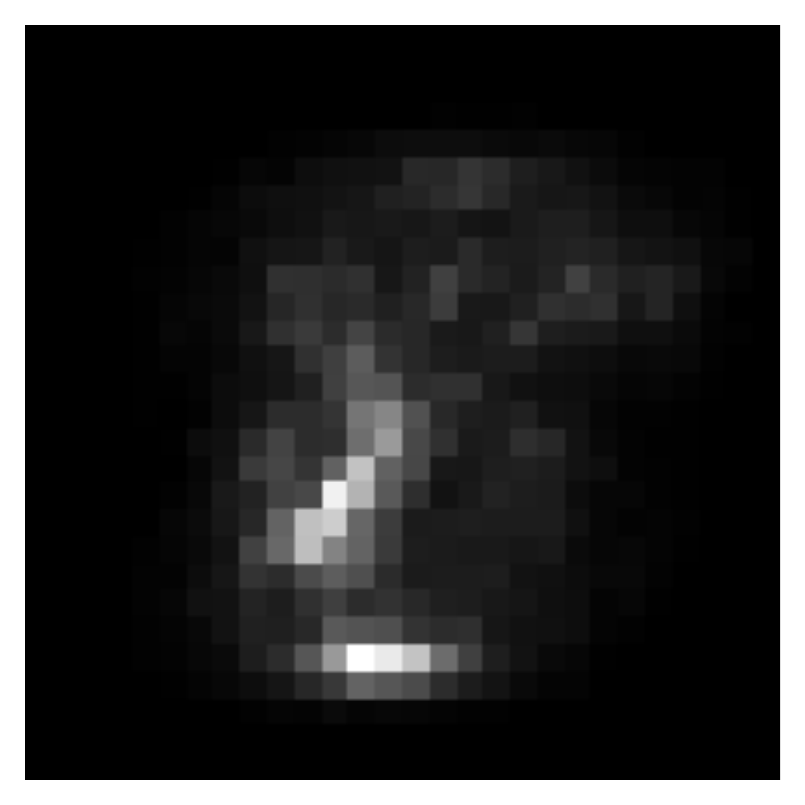}
    \end{minipage}
    \begin{minipage}{0.09\textwidth}
        \centering
        \includegraphics[scale = 0.15]{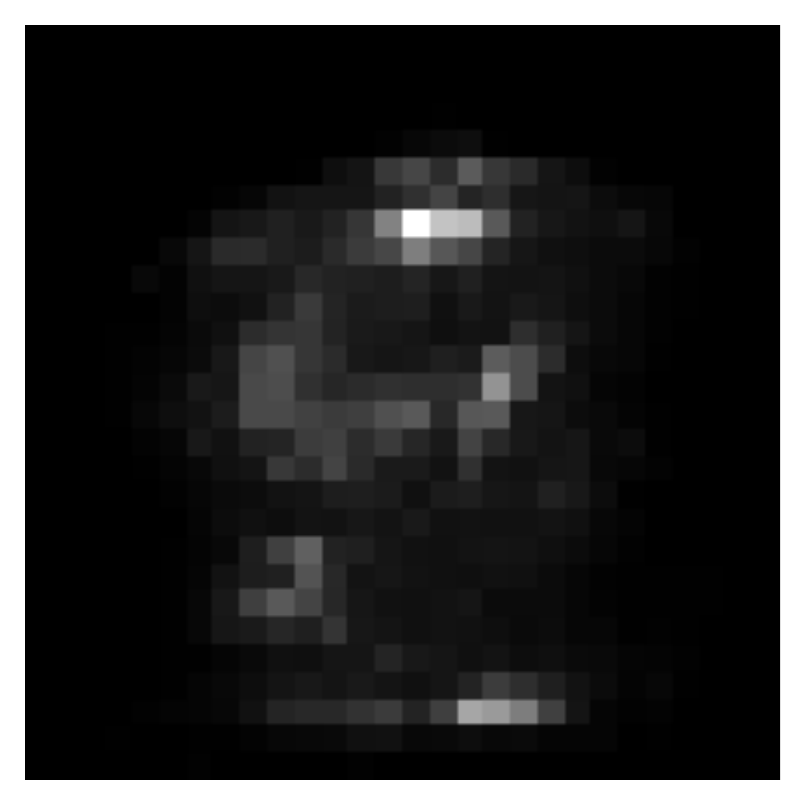}
    \end{minipage}
    
    \begin{minipage}{0.09\textwidth}
        \centering
        \includegraphics[scale = 0.15]{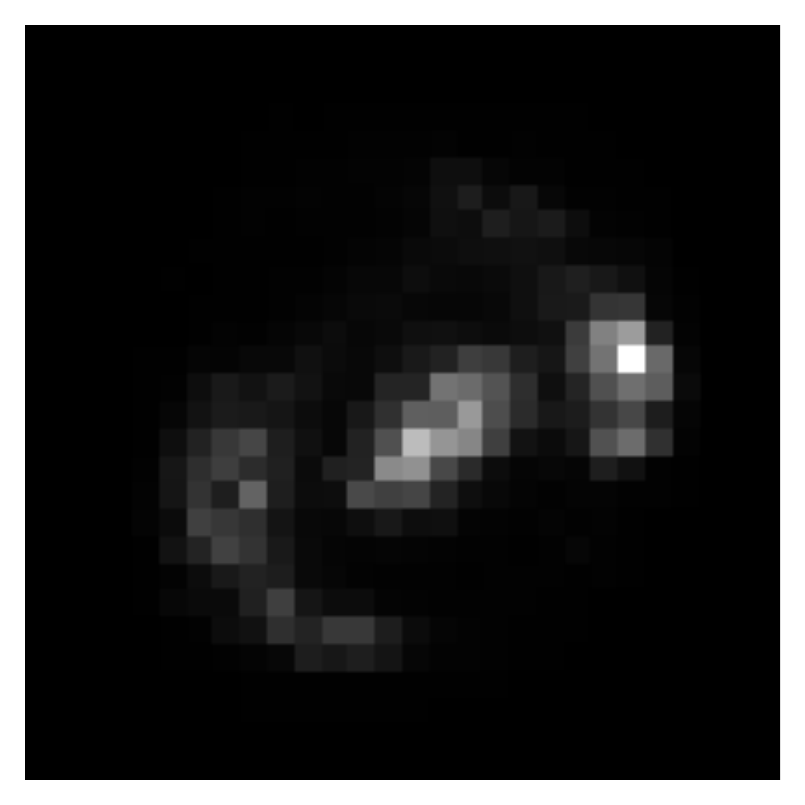}
    \end{minipage}
    \begin{minipage}{0.09\textwidth}
        \centering
        \includegraphics[scale = 0.15]{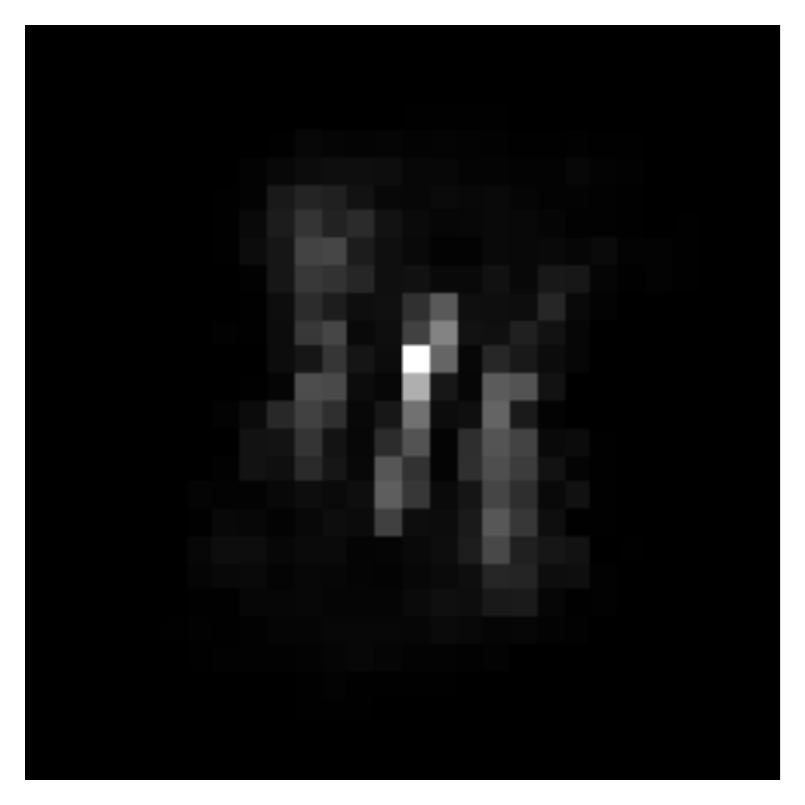}
    \end{minipage}
    \begin{minipage}{0.09\textwidth}
        \centering
        \includegraphics[scale = 0.15]{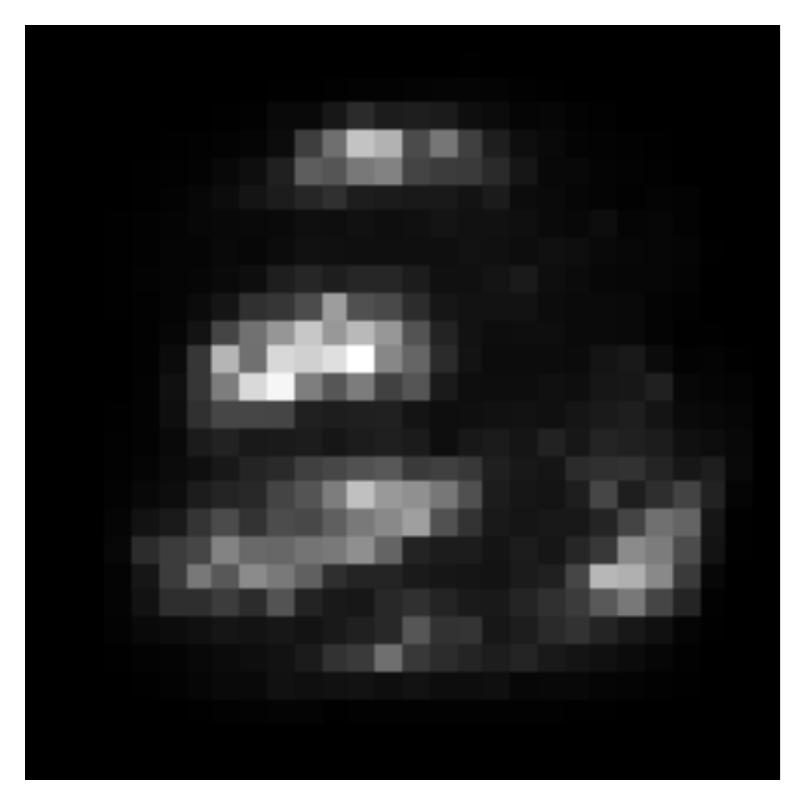}
    \end{minipage}
    \begin{minipage}{0.09\textwidth}
        \centering
        \includegraphics[scale = 0.15]{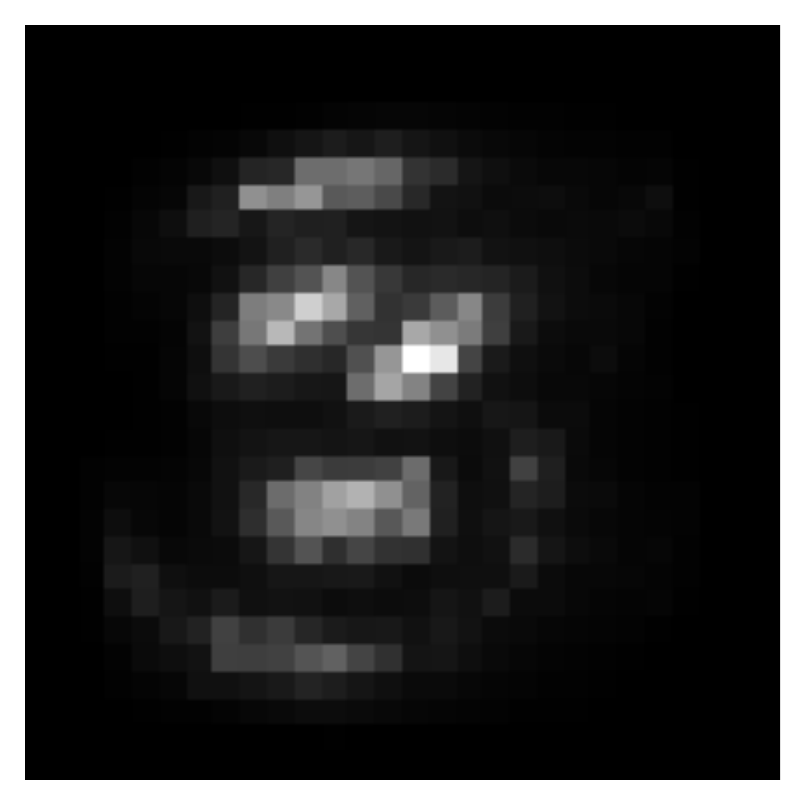}
    \end{minipage}
    \begin{minipage}{0.09\textwidth}
        \centering
        \includegraphics[scale = 0.15]{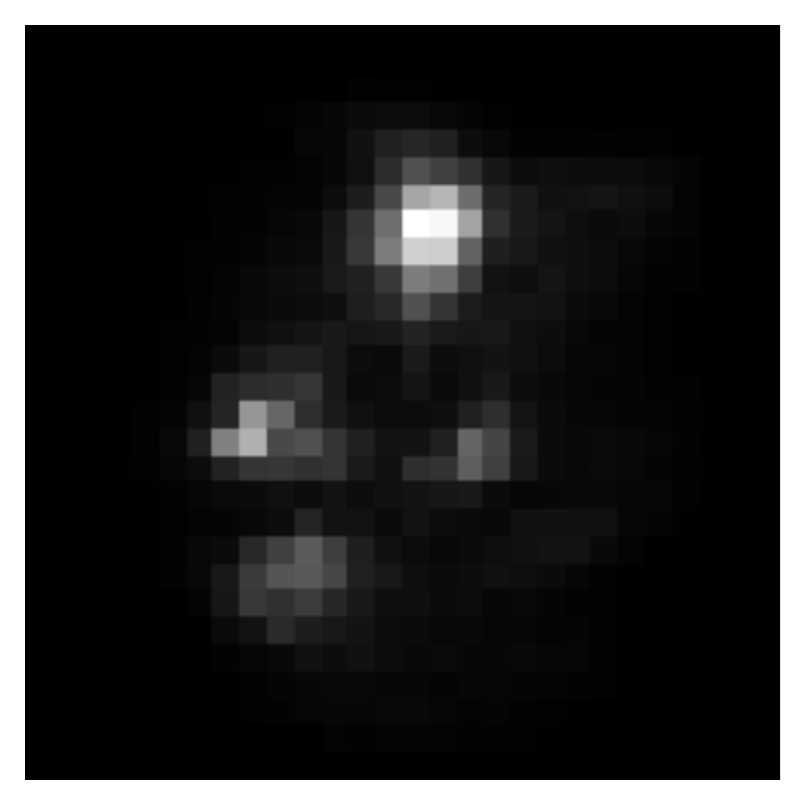}
    \end{minipage}
    \begin{minipage}{0.09\textwidth}
        \centering
        \includegraphics[scale = 0.15]{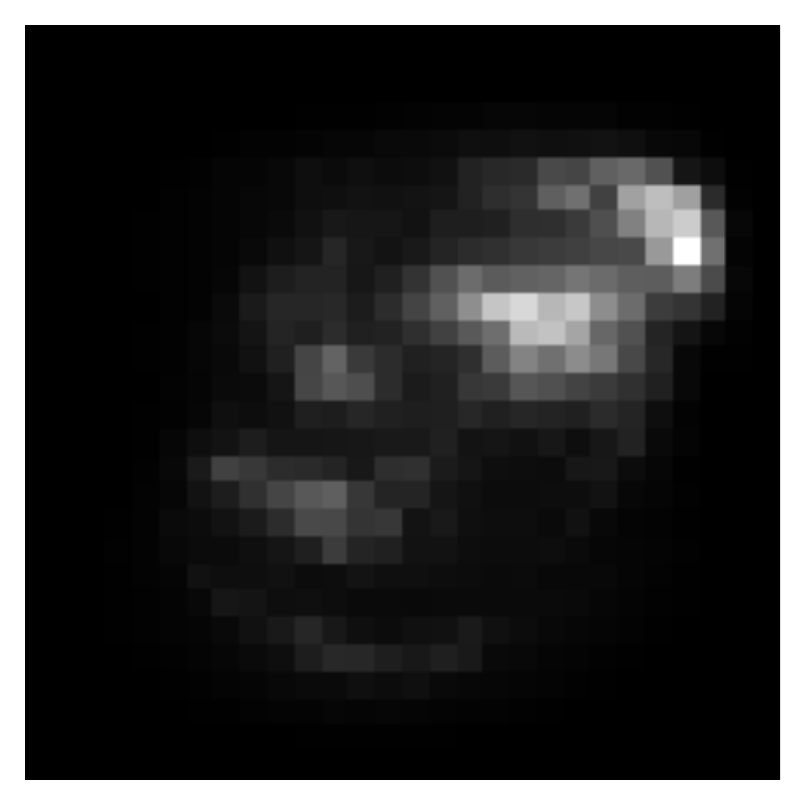}
    \end{minipage}
    \begin{minipage}{0.09\textwidth}
        \centering
        \includegraphics[scale = 0.15]{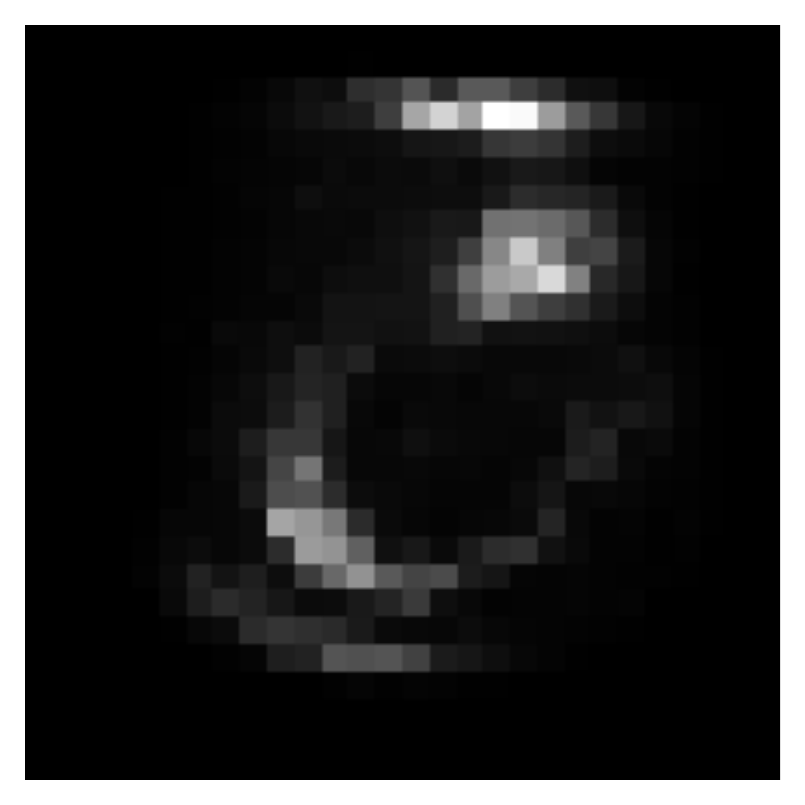}
    \end{minipage}
    \begin{minipage}{0.09\textwidth}
        \centering
        \includegraphics[scale = 0.15]{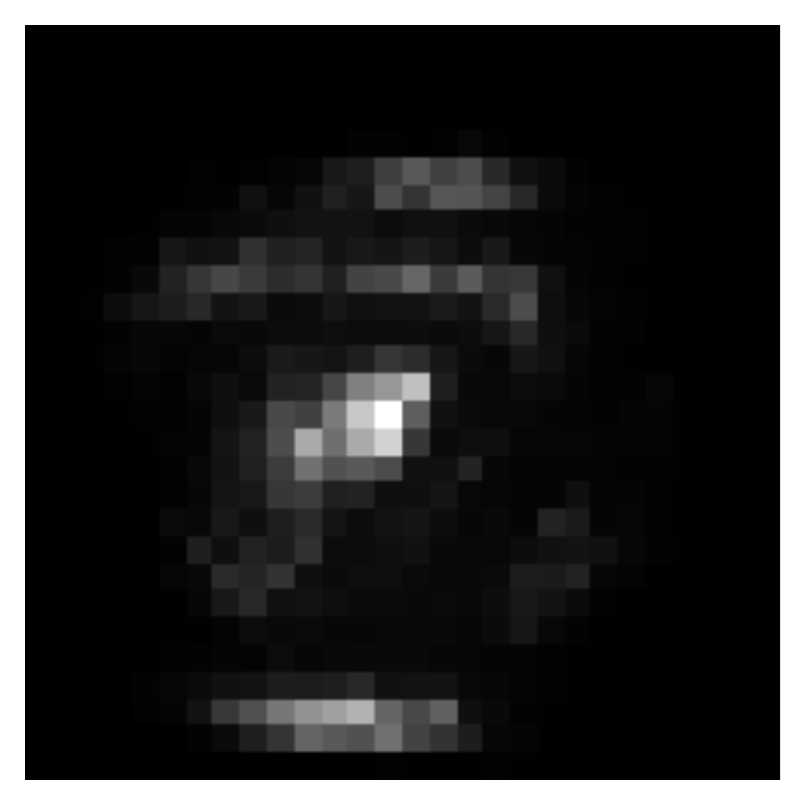}
    \end{minipage}
    \begin{minipage}{0.09\textwidth}
        \centering
        \includegraphics[scale = 0.15]{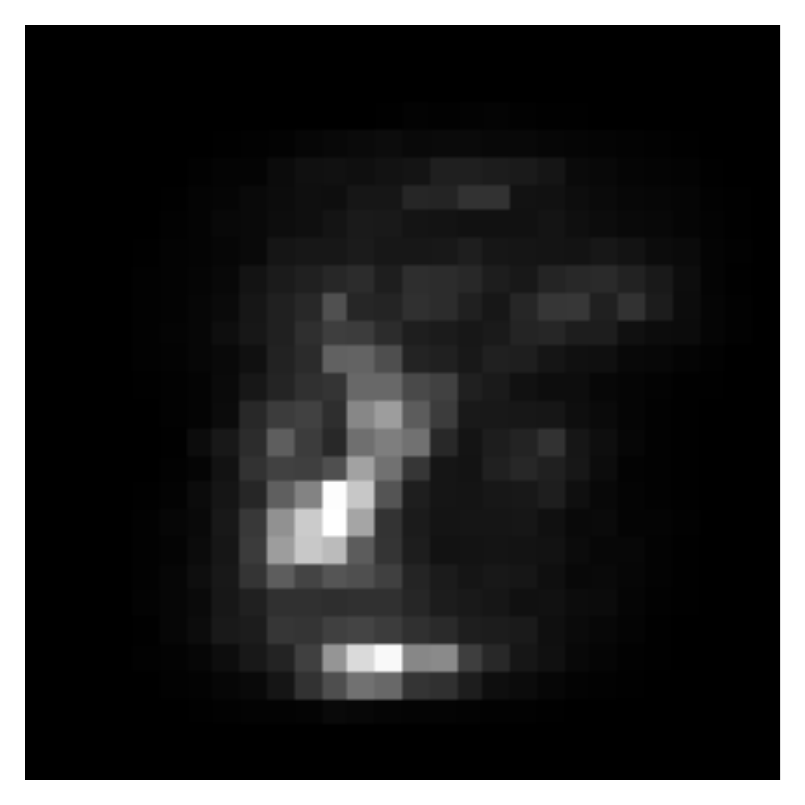}
    \end{minipage}
    \begin{minipage}{0.09\textwidth}
        \centering
        \includegraphics[scale = 0.15]{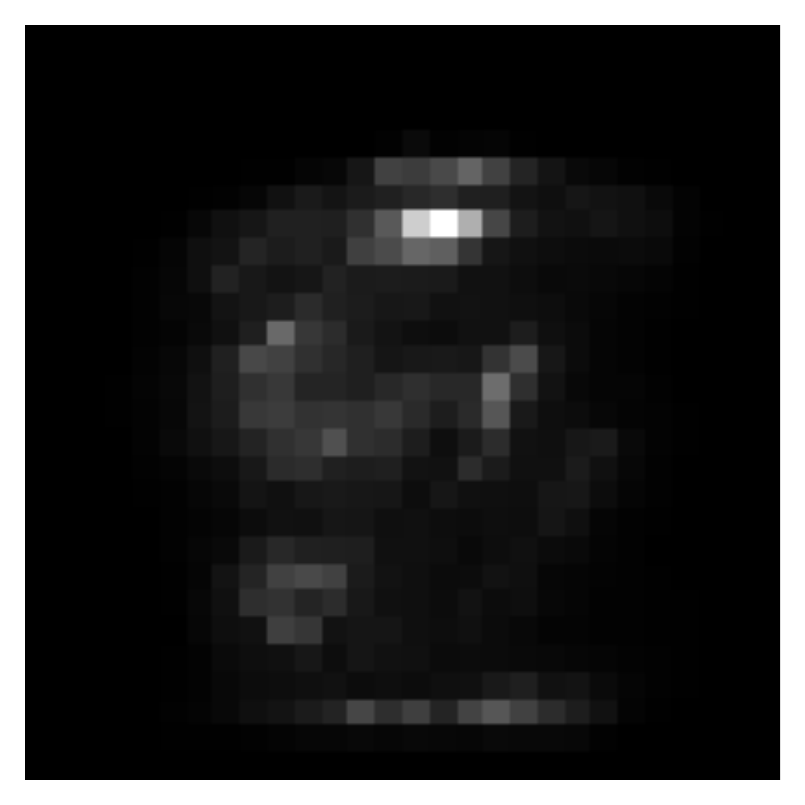}
    \end{minipage}
    \caption{Risk reduction importances learnt from ET classifier on MNIST digits. Top: PU ET with nonnegative risk estimator and quadratic loss. Bottom: PN ET with Gini impurity.}
    \label{fig:mnist_importances}
\end{figure}

Interpretability is a desirable property of machine learning algorithm, but 
neural networks are often hard to interpret. In contrast, tree based methods are generally easier to interpret.
We illustrate our risk reduction importances of both PU ET and PN
(positive-negative) ET on the UNSW-NB15 and the MNIST datasets (lighter color indicates a higher score) in \Cref{fig:unsw_importances} and
\Cref{fig:mnist_importances} respectively
.

%\raggedbottom

For MNIST, the plot of the importance scores for PU ET for each digit often
suggests the shape of the digit.
On the other hand, the Appendix provides some additional figures that show that
the normalized risk reduction importance makes many more pixels more important.
This observation is consistent with our discussion in \Cref{sec:puet}.
Interestingly, the importances for PU ET appear to be very similar
to those for PN ET (visually similar importance maps for MNIST, and 
qualitatively similar importance values which give similar feature rankings for
UNSW-NB15).
This suggests that the PU model is likely quite similar to the PN model.

\raggedbottom

\section{Related Work} \label{sec:related}
PU learning was initially studied in
\cite{denis1998pac,comite1999positive,letouzey2000learning}, and many approaches
have been developed based on different assumptions on the data generation
mechanism. 

The two-step approach discussed in the introduction have been adopted in many
papers with numerous instantiations.
We refer the readers to \cite{bekker2020learning} for a comprehensive list.

Another important line of work, which is most related to our work and has
achieved the state-of-the-art performance recently, minimizes a risk estimate
based on a weighted dataset of positive and negative examples obtained by
transforming the PU dataset
\cite{lee2003learning,elkan2008learning,du2014analysis,kiryo2017positive}.
Our work follows the risk minimization approach, but we develop general
risk minimization algorithms for learning tree models, that allows exploiting
the under-explored potential of random forests.

Numerous works have been done on tree-based methods, and random forest
algorithms are among the most successful ones
\cite{breiman2001random,geurts2006extremely,athey2016generalized}.
In the PU learning literature, PU decision trees were explored in early works
\cite{comite1999positive,letouzey2000learning}, and 
PU bagging \cite{mordelet2014bagging} has been explored for SVMs
\cite{mordelet2014bagging} and decision trees \cite{wu2021landslide}.
All of these approaches do not aim to directly minimize the risk.

Our recursive greedy risk minimization framework gives a new interpretation to
existing tree learning algorithms for positive and negative data, and allows us
to develop new algorithms for positive and unlabeled data.
In addition, our approach allows us to formulate a new feature importance score
that directly measures a feature's contribution to risk minimization.

\section{Conclusions}\label{sec:conclusions}
Random forests perform very well on fully supervised learning methods, but its
potential in PU learning has been under-explored due to the strong emphasis on
neural networks and the lack of a principled method for learning decision trees
from PU data.
We fill this gap in this paper. 
We first consider learning from positive and negative data, and develop a
recursive greedy risk minimization approach for learning decision trees, and
show that with suitable choice of loss functions, its instantiations are
equivalent to standard impurity-based decision tree learning algorithms.
We extend this approach to the PU setting, and develop a very efficient random
forest algorithm which requires little hyperparameter and yet strong performance
against the state-of-the-art PU learning algorithms.
Our approach supports a new feature importance score that directly measures a
feature's contribution to risk mnimization.

Our work opens up some new opportunities to build tree models.
First, decision trees can easily overfit the training set, and various
heuristics have been used in practice to alleviate overfitting.
Our recursive greedy risk minimization framework provides an alternative way to
control the size of the learned decision by always expanding the node with the
largest risk reduction until the risk reduction is smaller than a threshold or
when the tree size is larger than a threshold.
Second, we focused on quadratic loss and logistic loss in our experiments, but
we can also study alternative losses such as the hinge loss, or double hinge loss.

\begin{ack}
The authors would like to thank Dr Miao Xie from AARNet for his valuable suggestions. This work was funded by The University of Queensland Cyber Security Seed Funding. 
\end{ack}

%\bibliographystyle{plain}
%\bibliography{ref}

%%%%%%%%%%%%%%%%%%%%%%%%%%%%%%%%%%%%%%%%%%%%%%%%%%%%%%%%%%%%

\appendix
\pagebreak
%%%%%%%%%%%%%%%%%%%%%%%%%%%%%%%%%%%%%%%%%%%%%%%%%%%%%%%%%%%%%%%%%%%%%%%%%%%%%%%%%%%%%%%%%%%%%%%%%%%%%%%%
% \section{Proof of \Cref{thm:pnir}}\label{app:pnir}
\section{Proof of Theorem 1}\label{app:pnir}

\pnir
\begin{proof}
	(a) This result is proven in \cite{CART}. 
	\bigskip
	(b) By grouping the positive and negative examples, we have 
	\begin{align*}
		\hat{R}_{\Logistic}(v;S)  
		= \sum_{(\bx,y)\in S} w\ln(1+e^{-vy})
		= |S| w q_{+} \ln(1 + e^{-v}) + |S| w q_{-} \ln(1 + e^{v}).
	\end{align*}
	The minimizer satisfies 
	\begin{align*}
		\frac{\mathrm{d} \hat{R}_{\Logistic}}{\mathrm{d} v} 
		&= |S| w q_{+} \frac{-e^{-v}}{1 + e^{-v}} + |S| w q_{-} \frac{e^{v}}{1 + e^{v}}
		= 0.
	\end{align*}
	Solving the equation, we have 
	\begin{align*}
		v_S^* &= \ln(q_+/q_{-}),
	\end{align*}
	or $e^{v_{S}^{*}} = q_{+}/q_{-}$.
	Hence we have 
	\begin{align*}
		\hat{R}_{\Logistic}^*(S) 
		= |S| w q_{+} \ln\left(1 + \frac{q_{-}}{q_{+}}\right) + |S| w q_{-} \ln\left(1 + \frac{q_{+}}{q_{-}}\right)
		= |S|wH(S).
	\end{align*}
\end{proof}

% \section{Proof of \Cref{prop:uPUOptRisk}}\label{app:uPUOptRisk}
\section{Proof of Proposition 1}\label{app:uPUOptRisk}

\uPUOptRisk

Both the set $P' \subseteq P$ and the set $U' \subseteq U$ can possibly be
empty, thus $W_{\p} = |P'| w_{\p} \ge 0$,
$W_{\p} + W_{\n} = |U'| w_{\mathsf{u}} \ge 0$, and 
$W_{\n} = |U'| w_{\mathsf{u}} - |P'| w_{\p}$ can be negative, 0, or positive.
Note that we do not consider the case that $P'$ and $U'$ are both empty, as this
only happens when no examples are allocated to a node, which does not arise when
we construct decision trees.

\begin{proof}
	(a) For the quadratic loss, we have
	\begin{align*}
		\hat{R}_{\uPU} (v;P',U') 
		&= \sum_{x\in P'} w_\p (1-v)^2 - \sum_{x\in P'} w_\p (1+v)^2 + \sum_{x\in U'} w_\mathsf{u} (1+v)^2 \\
		&= W_\p (1-v)^2 + W_\n (1+v)^2.
	\end{align*}
	If $W_\p + W_\n \ne 0$, then the partial empirical risk is a convex quadratic
	function with a unique minimizer satisfying 
	\begin{align*} 
		\frac{\text{d} \hat{R}_{\uPU}}{\text{d} v} 
		= -2W_\p (1-v_{P',U'}^*) + 2W_\n  (1+v_{P',U'}^*) 
		= 0.
	\end{align*}
	Solving the equation, we have 
	\begin{align*}
		v_{P',U'}^* = 2v^{*} - 1.
	\end{align*}
	Therefore, we have
	\begin{align*}
		\hat{R}_{\uPU}^* (P',U') 
		&= (W_{\p} + W_{\n}) v^{*} \left( 1-2 v^{*} + 1 \right)^2 
			+ (W_{\p} + W_{\n}) (1-v^{*}) \left(1+ 2 v^{*} - 1 \right)^2 \\
		&= 4(W_\p + W_\n) v^*\left( 1-v^* \right). 
	\end{align*}
	If $W_{\p} + W_{\n} = 0$, then 
	$\hat{R}_{\uPU}(v; P', U') 
	= -2 W_\p v + W_{\p} + 2W_\n v + W_{\n}$.
	This gives
	$v_{P',U'}^* = +\infty$, and $\hat{R}_{\uPU}^* (P',U') = -\infty$. 

	\bigskip
	(b) For the logistic loss, we have
	\begin{align*}
		\hat{R}_{\uPU}(v;P',U') &= \sum_{\bx\in P'} w_\p \ln(1+ e^{-v}) - \sum_{\bx\in P'} w_\p \ln(1+e^{v}) + \sum_{\bx\in U'} w_{\mathsf{u}} \ln(1+e^{v}) 
		\\
		&= W_\p \ln(1 + e^{-v}) + W_\n \ln(1 + e^{v}).
	\end{align*}
	The derivative of the partial empirical risk wrt $v$ is 
	\begin{align*}
		\frac{\text{d}\hat{R}_{\uPU}}{\text{d} v} 
		&= \frac{W_\n e^{v} - W_\p}{1 + e^{v}}.
	\end{align*}
	By inspecting how the sign of the derivative changes as $v$ changes, we can
	obtain 
	\begin{align}
		v_{P',U'}^* = 
		\begin{cases}  
			\ln(W_\p/W_\n), & W_\n >0, W_\p > 0, \\
			-\infty, & W_\n >0, W_\p=0, \\
			+\infty, & W_{\n} \le 0.
		\end{cases} \label{eq:logistic_minimiser}
	\end{align}
	with
	\[
		\hat{R}_{\uPU}^*(P',U') = 
		\begin{cases}
			(W_\p + W_\n) (-v^* \ln(v^*) - (1-v^*)\ln(1-v^*)), & 0<v^*<1, \\
			0, & v^*\in\{0,1\},\\
			-\infty, & v^*>1. 
		\end{cases}
	\]
\end{proof}

\section{Unbiased Estimates of Proportions}\label{app:unbiased}
The weight $W_\p$ is an unbiased estimate of the probability $p(\bx\in\kappa, y=1)=\pi p(\bx\in\kappa\gvn y=1)$ since $\pi |P'|/|P|=|P'|w_\p=W_\p$ estimates the probability $\pi p(\bx\in\kappa\gvn y=1)$ without bias.

To show that $W_\n$ is an unbiased estimate of the probability $p(\bx\in\kappa, y=-1)$, first note that for any $\bx$ we have $(1-\pi) p_\n(\bx) = p(\bx)-\pi p_\p(\bx)$. Integrating over $\bx\in\kappa$, we have $(1-\pi)p(\bx\in\kappa\gvn y=-1) = p(\bx\in\kappa) - \pi p(\bx\in\kappa\gvn y=1)$, or equivalently, $p(\bx\in\kappa,y=-1) = p(\bx\in\kappa) - \pi p(\bx\in\kappa\gvn y=1)$. Now, $p(\bx\in\kappa)$ is unbiasedly estimated by $|U'|w_\mathsf{u}$, and $\pi p(\bx\in\kappa\gvn y=1)$ is unbiasedly estimated by $|P'|w_\p$.

% \section{Proof of \Cref{prop:nnPUOptRisk}}
\section{Proof of Proposition 2}
\nnPUOptRisk
\begin{proof}
To find the minimal partial \nnPU empirical risk, we begin by computing the derivative of $\hat{R}_{\nnPU}$ with respect to $v$:
\begin{align*}
    \hat{R}_{\nnPU} (v;P',U') &= \sum_{\bx\in P'} w_\p (1-v)^2  + \max\left\{0,\sum_{\bx\in U'} w_\mathsf{u} (1+v)^2 - \sum_{\bx\in P'} w_\p (1+v)^2  \right\}
    \\
    &= W_\p (1-v)^2 + \max\{0,W_\n (1+v)^2\},
    \\
    \frac{\text{d} \hat{R}_{\nnPU}}{\text{d} v} &= 2W_\p (v-1) + 2\max\{0,W_\n\} (v+1).
\end{align*}
First consider the case when $v^{*} > 1$, then either $\frac{W_\p}{W_\p+W_\n} > 1$ or $v^{*} = +\infty$. The former suggests that $W_\n < 0$, and hence the minimizer will be 1. In the latter we have $W_\n = -W_\p \leq 0$, again giving us a minimizer of 1. On the other hand if $0\leq v^*<1$, then we have $W_\p\ge 0$ and $W_\n > 0$ and the minimizer will be the solution to $0=\text{d} \hat{R}_{\nnPU}/\text{d} v$.

The minimizer can thus be written succinctly as 
\[
    \underset{v\in\R}{\argmin}\,\hat{R}_{\nnPU} (v;P',U') = 
    \begin{cases}
        1,& v^* > 1, \\
				2v^{*} - 1,& \text{otherwise},
    \end{cases}
\]
with minimum partial empirical risk
\[
    \hat{R}_{\nnPU}^* (v;P',U') = 
    \begin{cases}
        0,& v^*>1,\\
        4(W_\p + W_\n) v^*(1-v^*), & \mathrm{otherwise}.
    \end{cases}
\]
For the logistic loss we shall again start by computing the derivative of the partial empirical risk:
\begin{align*}
    \hat{R}_{\nnPU} (v;P',U') 
    &= W_\p \ln(1+e^{-v}) + \max\{0,W_\n \ln(1+e^{v})\}
    \\
    \frac{\text{d}\hat{R}_{\nnPU} }{\text{d} v} &= \frac{\max\{0,W_\n\}e^{v} - W_\p}{1+e^{v}}.
\end{align*}
In this case, the minimizer $\argmin_{v\in\R}\hat{R}_{\nnPU}(v;P',U')$ is the same as in \eqref{eq:logistic_minimiser}. The corresponding minimum partial risk is 
\[
    \hat{R}_{\nnPU}^*(P',U') = 
    \begin{cases}
        (W_\p+W_\n)(-v^*\ln(v^*) - (1-v^*)\ln(1-v^*)), & 0<v^*<1, \\
        0, & \text{otherwise}.
    \end{cases}
\]
\end{proof}

\section{Optimizing Split Time Complexity}\label{app:optimising_split}
Algorithm \ref{alg:optimal_split} gives pseudocode for finding the optimal split for a given feature. 
Each step is annotated with its time complexity, where $n = |P'| + |U'|$.
Note that computing $\hat{R}^{*}$ value can be done in constant time if $W_{\p}$
and $W_{\n}$ values are given.
The time complexity of Algorithm \ref{alg:optimal_split} is $O((|P'|+|U'|)\ln(|P'|+|U'|) + m)$, with $m\leq |P'|+|U'|+1$.
\begin{algorithm}
\KwIn{Feature $f$, set $P'$ of positive examples and set $U'$ of unlabeled examples.}
\KwOut{Split $(f,t)$ that gives the largest risk reduction.}

     Sort all examples $\bx_1,\ldots,\bx_n$ in $P'\cup U'$ by their entries in feature $f$. 
		 Denote the sorted examples by $\bx_{(1)}, \ldots, \bx_{(n)}$ \Comment*[r]{$O(n\ln n)$}
		 Let the distinct $f$ values be $u_{1} < u_{2} < \ldots < u_{m+1}$, and choose 
		 $t_{i} = (u_{i} + u_{i+1})/2$, $i = 1, \ldots, m$. Note $m+1\leq n$ \Comment*[r]{$O(m)$} 
		Compute the $\hat{R}^{*}(P', U')$ value by computing its $W_{\p}$ and $W_{\n}$ values \Comment*[r]{$O(n)$}
		Compute the risk reduction for $(f,t_1)$ by computing the $W_{\p}$ and
		$W_{\n}$ values for the child nodes \Comment*[r]{$O(n)$}
    $t^*\gets t_1$ \Comment*[r]{$O(1)$}
		 \For(\tcp*[h]{$O(n)$ in total}){$i=2,\ldots,m$}{
			 Compute the risk reduction for $(f,t_i)$. This only requires updating
			 $W_{\p}$ and $W_{\n}$ values for $t_{i-1}$ by considering those
			 $\bx_{(i)}$ with $f$ values between $t_{i-1}$ and $t_{i}$\;
     \If{Risk reduction for $(f,t_i)>$ risk reduction for $(f,t^*)$}
		 {$t^*\gets t_i$\;} %\Comment*[r]{$O(1)$}}
     }  
\textbf{Return}: $t^*$.
\caption{\texttt{Find Optimal Threshold}}
\label{alg:optimal_split}
\end{algorithm}

% \section{Proof of \Cref{prop:quadsavage}}
\section{Proof of Proposition 3}
In addition, we show that the sigmoid loss, $\hat{R}^{*}_{\uPU}(P', U')$ has an interesting interpretation.

\quadsavage
\begin{proof}
	We have already shown that for the quadratic loss, 
	$\hat{R}^{*}(S) = 2 |S| w G(S) = 4 |S| w q_{+} q_{-}$.

	For the savage loss, we have
	\begin{align*}
		\hat{R}(v; S) 
		= \sum_{(x, y)} w \frac{4}{(1 + e^{v y})^{2}}
		= |S| w \left[q_{+} \frac{4}{(1 + e^{v})^{2}} + q_{-} \frac{4}{(1 + e^{-v})^{2}}\right].
	\end{align*}
	The derivative of $\hat{R}(v; S)$ wrt $v$ is 
	\begin{align*}
		\frac{\text{d} \hat{R}}{\text{d} v} 
		=
		8|S| w \frac{-q_{+} e^{v} + q_{-} e^{2v}}{(1 + e^{v})^{3}}.
	\end{align*}
	Setting the derivative to 0, we obtain $v^{*}_{S} = \ln \frac{q_{+}}{q_{-}}$.
	Therefore
	\begin{align*}
		\hat{R}^{*}(S)
		= |S| w \left[q_{+} \frac{4}{(1 + q_{+}/q_{-})^{2}} + q_{-} \frac{4}{(1 + q_{-}/q_{+})^{2}}\right]
		= 4|S| w q_{+} q_{-},
	\end{align*}
	which is the same as the $\hat{R}^{*}(S)$ value for the quadratic loss.
\end{proof}

\addtocounter{proposition}{4}
\begin{proposition}
	For the sigmoid loss, we have
	\begin{align}
			\hat{R}_{\uPU}^*(P',U') 
			= 
			\min\{W_{\p}, W_{\n}\}.
			\label{eq:sigmoid_impurity}
	\end{align}
\end{proposition}
Intuitively, the above expression uses the weight of the rarer class as the
impurity measure, and we can derive a similar expression in the PN setting too.
Such an impurity measure is an interesting one that has not been used in
decision tree learning.
\begin{proof}
First note that
\begin{align*}
    \hat{R}_{\uPU}(v;P',U') &= \sum_{\bx\in P'} \frac{w_\p}{1+e^{v}} - \sum_{\bx\in P'} \frac{w_\p}{1+e^{-v}} + \sum_{\bx\in U'} \frac{w_\mathsf{u}}{1+e^{-v}}\nonumber
    \\
    &= \frac{W_\p}{1+e^{v}} + \frac{W_\n}{1+e^{-v}} \nonumber
\end{align*}
The derivative of the partial empirical risk is
\begin{align}
    \frac{\text{d}\hat{R}_{\uPU}}{\text{d} v} &= \frac{e^{v}(W_\n-W_\p)}{(1+e^{v})^2}.\label{eq:sig_deriv}
\end{align}
The equation \eqref{eq:sig_deriv} tells us that there are no stationary points in the partial empirical risk. The minimizer $v_{P',U'}^*$ is given by:
\begin{align*}
    v_{P',U'}^* 
		= 
    \begin{cases}
        +\infty, &  W_\p > W_\n,\\
				\text{any value}, & W_{\p} = W_{\n}, \\
        -\infty, &  W_\p < W_\n,\\
    \end{cases}
    = 
    \begin{cases}
        +\infty, &  v^* > 0.5,\\
				\text{any value}, & v^{*} = 0.5, \\
        -\infty, &  v^* < 0.5,\\
    \end{cases}
\end{align*}
The corresponding minimum partial empirical risk is given by
\begin{align}
    \hat{R}_{\uPU}^*(P',U') = 
    \begin{cases}
        W_\n,& v^* > 0.5, \\
        W_\p,& v^* \leqslant 0.5.   
    \end{cases}
		=
		\min\{W_{\p}, W_{\n}\}.
\end{align}
\end{proof}

Note that we can derive similar expressions for each loss function described in
\Cref{sec:background}, however in this paper we only focus on the quadratic and
logistic loss due to their connection with the traditional Gini and entropy
impurities, respectively.

% \section{Proof of \Cref{prop:optimal_prediction_function}}
\section{Proof of Proposition 4}
\optimalpredictionfunction
\begin{proof}
For each loss function mentioned in \Cref{sec:background} we have $\ell(v,y)=a$ if $v=y$ and $\ell(v,y)=b$ if $v\ne y$, for some $0\leq a<b$ (with $a,b$ not necessarily common to all loss functions). The optimal prediction based on the \uPU risk estimator is $+1$ if $\hat{R}_{\uPU}(+1;P',U') < \hat{R}_{\uPU}(-1;P',U')$, that is, if
\begin{align*}
    W_\p \ell(+1,+1) + W_\n \ell(+1,-1) &< W_\p \ell(-1,+1) + W_\n \ell(-1,-1)
    \\
    W_\p a + W_\n b &< W_\p b + W_\n a
    \\
    (W_\p+W_\n)/2 &< W_\p,
\end{align*}
which happens if $v^*>0.5$. A symmetric argument can be applied to show that the optimal prediction is $-1$ when $v^* < 0.5$. Continuing, observe that $a,b\geq 0$, so for the \nnPU risk estimator the optimal prediction is $+1$ if
\begin{align*}
    W_\p a + \max\{0,W_\n b\} &< W_\p b + \max\{0,W_\n a\}
    \\
    (b-a) \max\{0,W_\n\} &< W_\p (b-a)
    \\
    W_\n\leq \max\{0,W_\n\} &< W_\p
    \\
    W_\n &< W_\p,
\end{align*}
which is the same result as for the \uPU risk estimator. A symmetric argument shows that the optimal constant prediction based on the \nnPU risk estimator is $-1$ if $v^*<0.5$, and hence (if we predict $-1$ when $v^*=0.5$ ),
\begin{align*}
    2\I(v^*>0.5)-1 
% 		&= \argmin_{v\in\{-1,+1\}}\,\hat{R}_{\uPU}(v;P',U') 
        &= \argmin_{v\in\{-1,+1\}}\,\hat{R}_{\nnPU}(v;P',U').
\end{align*}
% for all loss functions mentioned in \Cref{sec:background}.
We stress that this result holds for any loss function $\ell$ satisfying $\ell(v,y)>\ell(y,y)\geq 0$, with $v\neq y$. Many common loss functions satisfy this condition, including all loss functions mentioned in \Cref{sec:background} as well as the hinge loss $\ell(v,y) = \max\{0,1-vy\}$, the double hinge loss $\ell(v,y) = \max\{0,(1-vy)/2,-vy\}$, the zero-one loss $\ell(v,y) = (1-\mathrm{sign}(vy))/2$, the ramp loss $\ell(v,y) = \max\{0,\min\{1,(1-vy)/2\}\}$ and the exponential loss $\ell(v,y) = \exp(-vy)$.
\end{proof}

\section{Pseudocode for PU ET}\label{app:pseudocode}
Our PU Extra Trees algorithm is shown in \Cref{alg:PUET}, with the learning
algorithm for each individual decision tree given in
\Cref{alg:construct_subtree}, and the method used to find a split for a node in
\Cref{alg:splitting}. 
A nice property of this algorithm is that each tree in the forest can be trained in parallel with no need to share data between processes. 
Notice the similarity between \Cref{alg:construct_subtree} and decision tree learning in the PN setting seen in \Cref{alg:learndt}, with the differences being our new method for computing the prediction value at a specific node and the sub-routine for choosing a split for a node as in 
\Cref{alg:splitting}.

To construct a random forest with 100 decision trees using the randomisation trick in Extra Trees \cite{geurts2006extremely}, one can run \texttt{PUET}$(100,P,U,\lceil \sqrt{d} \rceil,1)$.

\begin{algorithm}
% \SetAlgoLined
\KwIn{$N$ - number of trees; $P$ - positive examples; $U$ - unlabeled examples;
	$F$ - number of attributes; $T$ - number of threshold values per attribute.}
\KwOut{Collection of trained decision trees.}
    Initialise collection of decision trees $\mathcal{T}\gets \emptyset$\;
    \For{$i=1,\ldots,N$}{
    Initialise the root node $\kappa_i$ for decision tree $i$\;
    Train a single decision tree $\mathbb{T}$ using $\texttt{Construct\_Subtree}(\kappa_i, P,U,F,T)$\;
    $\mathcal{T}\gets \mathcal{T}\cup \mathbb{T}$\;
    }
    
\textbf{Return}: $\mathcal{T}$
	\caption{\texttt{PUET}$(N, P, U, F, T)$}
\label{alg:PUET}
\end{algorithm}

\begin{algorithm}
\SetAlgoLined
\KwIn{$\kappa$ - node; 
	$P'$ - positive examples at $\kappa$; 
	$U'$ - unlabeled examples at $\kappa$; 
	$F$ -number of attributes; 
	$T$ - number of threshold values per attribute.} 
\KwOut{A (sub) decision tree $\mathbb{T}_\kappa$.}
\eIf{termination\ criterion\ is\ met}{
    Compute optimal prediction value $v_{P',U'}^*=\argmin_{v\in\{-1,+1\}}\,\hat{R}_{\nnPU}(v;P',U')$ for node $\kappa$\;
    }{
    Choose a split $(f,t)$ for node $\kappa$ using $\texttt{Find\_Split}(\kappa, F,T)$\;
    Create two child nodes $\kappa_{f>t}$ and $\kappa_{f\leq t}$ for $\kappa$ and split data into $P'_{f > t}$, $P'_{f \leq t}$, $U'_{f > t}$, $U'_{f\leq t}$\;
    $\mathbb{T}_{\kappa_{f>t}}\leftarrow $ \texttt{Construct\_Subtree}$(\kappa_{f>t},\, P'_{f>t},\, U'_{f>t},F,T)$\;
    $\mathbb{T}_{\kappa_{f\leq t}}\leftarrow $ \texttt{Construct\_Subtree}$(\kappa_{f\leq t},\, P'_{f\leq t}, U'_{f\leq t},F,T)$\; 
    }
\textbf{Return}: $\mathbb{T}_\kappa$
\caption{\texttt{Construct\_Subtree}$(\kappa, P', U', F, T)$}
	\label{alg:construct_subtree}
\end{algorithm}

\begin{algorithm}
\KwIn{$\kappa$ - node; $F$ - number of attributes; $T$ - number of threshold values per attribute.}
\KwOut{Optimal split $(f,t)$.}
    Initialise collection of splits $\mathcal{S}\gets \emptyset$\;
    Select $F$ attributes $\{f\}$ uniformly at random from all non-constant attributes at node $\kappa$\;
    \For{each $\{f\}$}{
        \For{$i = 1,\ldots, T$}
            {Sample a random cut-point $t$ uniformly from the range of attribute $f$ at node $\kappa$\;
            $\mathcal{S}\gets \mathcal{S}\cup \{(f,t)\}$\;
            }
        }
    Choose splitting point $(f_\kappa,t_\kappa)\in\mathcal{S}$ that maximises the PU-data based risk reduction $	\hat{R}^{*}(P', U') - \hat{R}^{*}(P'_{f > t}, U'_{f > t}) - \hat{R}^{*}(P'_{f \le t}, U'_{f \le t})$ for node $k$\; 
\textbf{Return}: $(f,t)$.
	\caption{\texttt{Find\_Split}$(\kappa, F, T)$}
\label{alg:splitting}
\end{algorithm}

\section{Computational Efficiency of PU ET}\label{app:times}
We have run additional experiments to support our claim that PU ET is efficient. Our current Python implementation of PU ET (nnPU risk with quadratic loss) takes seconds to train a single-tree random forest on modest hardware (Intel Core i7-10700 CPU and 32 GB RAM) as shown in \Cref{tab:times}. Thus training a random forest is typically must faster than training a neural net.

\begin{table}[H]
    \centering
    \caption{Training time mean (sd) in seconds for a single tree random forest.}
    \begin{tabular}{lccc}
        Dataset & PU ET & PN ET (ours) & PN ET (sklearn) \\ \hline
        20news & 1.26 (0.03) & 5.39 (0.09) & 0.03 (0.01) \\
        Mushroom & 0.14 (0.02) & 0.07 (0.01) & 0.01 (0.01) \\
        MNIST & 5.90 (0.33) & 37.90 (0.89) & 0.34 (0.01) \\
        CIFAR-10 & 16.32 (0.36) & 119.78 (1.58) & 0.97 (0.01) \\
        UNSW-NB15 & 3.35 (0.38) & 67.60 (1.48) & 0.11 (0.01)
    \end{tabular}
    \label{tab:times}
\end{table}
A Cython implementation of PU ET will be much faster: sklearn's cython based implementation of PN ET (with Gini impurity) often achieves a speedup of two orders of magnitude as compared to our PN ET (with Gini impurity), on fully labeled datasets, as shown in \Cref{tab:times}. Note that our PN ET implementation is the same as our PU ET implementation, except different risk measures are used in the calculation of the splitting points.

\section{Hyperparameter Tuning for PU ET}\label{app:hyperparams}
We deliberately chose to not perform hyperparameter tuning for results in the main text to demonstrate that PU ET can be reasonably effective even without significant hyperparameter tuning. We have run some additional experiments to show that the default choice of hyperparameters offer strong performance with the benefit of decreased training times. We trained PU ET with \nnPU risk estimator and quadratic loss by varying $F$ (number of sampled features) and $T$ (number of thresholds sampled for each feature) to see their effect on training time and predictive performance. Results are given in \Cref{tab:hyperparam_accs} and \Cref{tab:hyperparam_times}. All experiments here were performed on an Intel Core i7-10700 CPU with 32 GB RAM using MNIST digits, 5 replications, 100 trees in the forest and under the usual PU learning experimental setup $(|P| = 1000, |U|=n)$. 
% Accuracies are shown below.
\begin{table}[!h]
    \centering
    \caption{Accuracy mean\% (sd) on the test set for PU ET using various hyperparameter combinations.}
    \begin{tabular}{ccc}
        $F$ & $T=10$ & $T=1$ \\ \hline
        $d$ & 89.75 (0.71) & 92.47 (0.54) \\
        $\lceil\sqrt{d} \rceil$ & 93.24 (0.28) & 93.74 (0.75) \\
        $1$ & 90.02 (0.68) & 81.03 (0.09) 
    \end{tabular}
    \label{tab:hyperparam_accs}
\end{table}
% Running times are shown below.
\begin{table}[!h]
    \centering
    \caption{Training time mean (sd) in seconds for PU ET using various hyperparameter combinations.}
    \begin{tabular}{ccc}
        $F$ & $T=10$ & $T=1$ \\ \hline
        $d$ & 15071.88 (172.07) & 1771.92 (31.24) \\
        $\lceil\sqrt{d} \rceil$ & 1231.89 (26.64) & 225.65 (1.65) \\
        $1$ & 370.54 (1.22) & 315.42 (5.58)
    \end{tabular}
    \label{tab:hyperparam_times}
\end{table}

The default setting $(F=\lceil\sqrt{d} \rceil, T=1)$ results in lower training times and relatively strong predictive performance. This matches the findings in the original ET paper \cite{geurts2006extremely} where this default was able to strike a good balance between training time and predictive performance.

\section{Overfitting with uPU Risk Estimators in PU ET}\label{app:overfitting}
We performed additional experiments to empirically investigate the difference between \uPU and \nnPU risk estimators in regards to overfitting. In \Cref{tab:overfitting} we report the training risks (measured as PU risk as data is PU) and testing risks (measured as PN risk as data is PN) using zero-one loss $\ell_{0/1}(v, y) = (1 - \mathrm{sign}(vy))/2$ on a number of datasets. Each result is reported as mean (sd) over 5 replications. From the results we can see that the training risk is significantly smaller than the test risk in the uPU setting as compared to the nnPU setting, confirming that uPU suffers more from overfitting than nnPU.

\begin{table}[!h]
    \centering
    \caption{Training and testing risk of PU ET. }
    \begin{tabular}{lccccc}
 	 	\multirow{2}{*}{Dataset} && \multicolumn{2}{c}{uPU} & \multicolumn{2}{c}{nnPU}	 \\ \cline{3-6}
 	 	            & & Quadratic & Logistic & Quadratic & Logistic \\ \hline
        \multirow{2}{*}{20News} & Train & -0.48 (0.00) & -0.22 (0.01) & 0.00 (0.00) & 0.00 (0.00) \\
         	 & Test & 0.56 (0.00) & 0.28 (0.01) & 0.15 (0.00) & 0.18 (0.01) \\ \hline
        \multirow{2}{*}{Mushroom} & Train & -0.36 (0.00) & -0.01 (0.00) & 0.00 (0.00) & 0.00 (0.00) \\
         	& Test & 0.39 (0.01) & 0.01 (0.00) & 0.00 (0.00) & 0.00 (0.00) \\\hline
        \multirow{2}{*}{MNIST} & Train & -0.47 (0.00) & -0.08 (0.01) & 0.00 (0.00) & 0.00 (0.00) \\
         	 & Test & 0.49 (0.00) & 0.11 (0.01) & 0.04 (0.00) & 0.06 (0.00) \\\hline
        \multirow{2}{*}{CIFAR-10} & Train & -0.38 (0.00) & -0.17 (0.01) & 0.00 (0.00) & 0.00 (0.00) \\
         	 & Test & 0.4 (0.00) & 0.25 (0.00) & 0.16 (0.00) & 0.21 (0.00) \\\hline
        \multirow{2}{*}{UNSW-NB15} & Train & -0.58 (0.00) & -0.03 (0.01) & 0.02 (0.01) & 0.02 (0.01) \\
         	 & Test & 0.65 (0.00) & 0.13 (0.01) & 0.1 (0.01) & 0.14 (0.00) 
    \end{tabular}
    \label{tab:overfitting}
\end{table}

\section{MNIST Feature Importances}
\begin{figure}[H]
    \centering
    \begin{minipage}{0.09\textwidth}
        \centering
        \includegraphics[scale = 0.15]{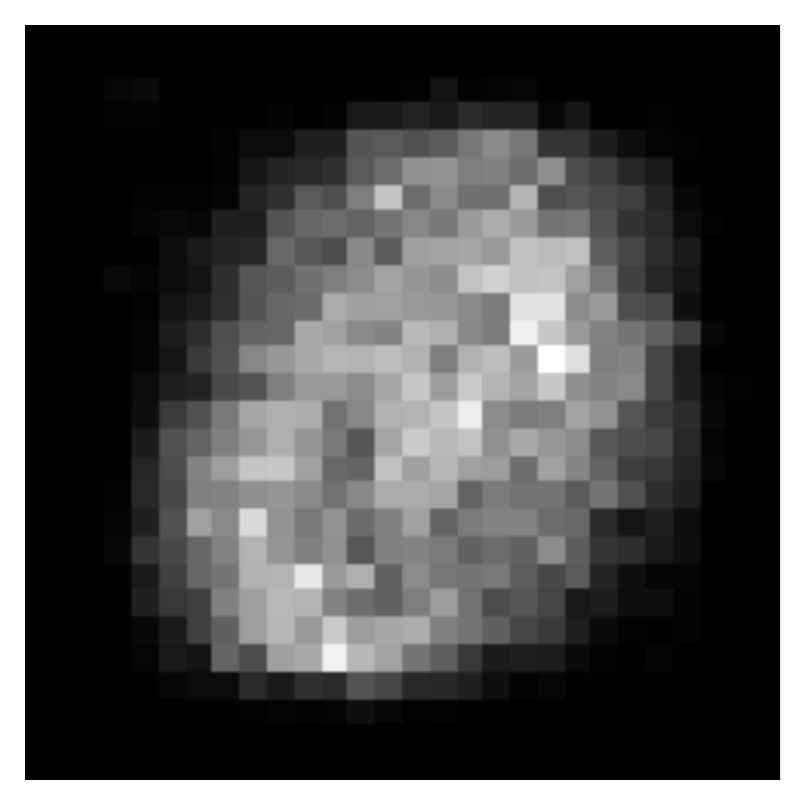}
    \end{minipage}
    \begin{minipage}{0.09\textwidth}
        \centering
        \includegraphics[scale = 0.15]{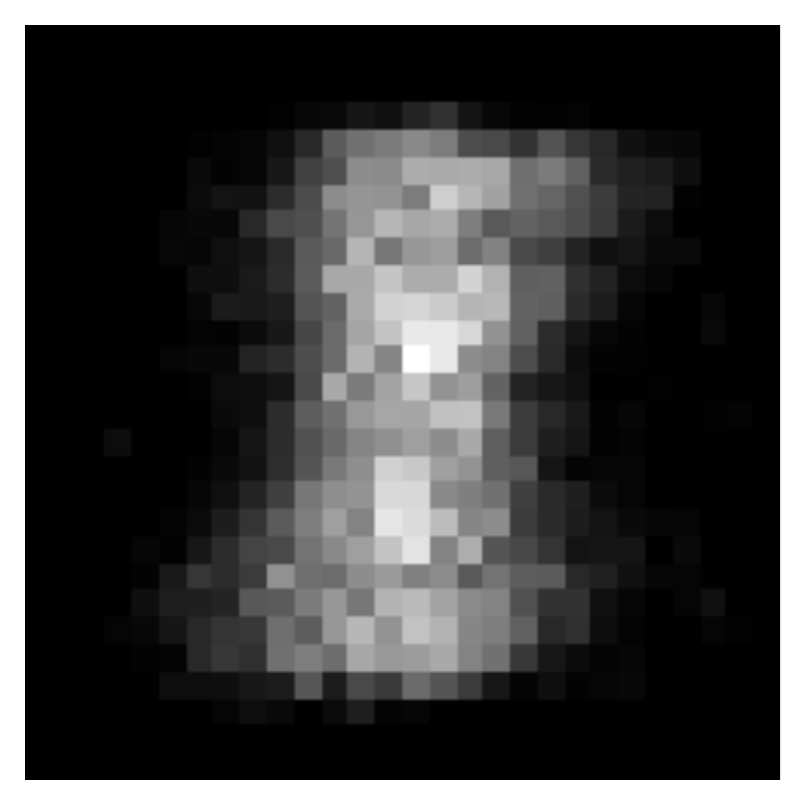}
    \end{minipage}
    \begin{minipage}{0.09\textwidth}
        \centering
        \includegraphics[scale = 0.15]{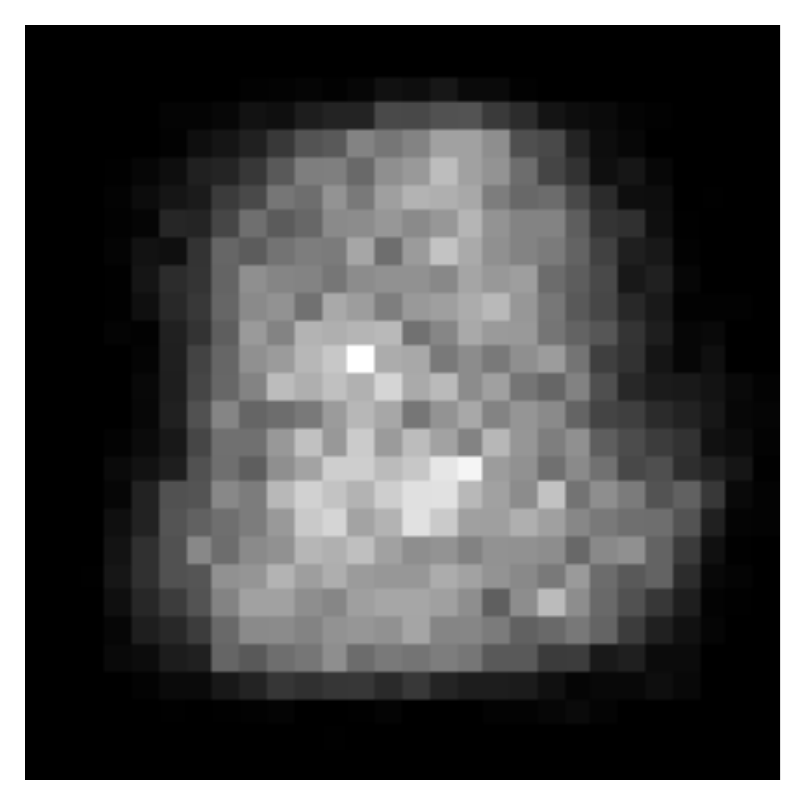}
    \end{minipage}
    \begin{minipage}{0.09\textwidth}
        \centering
        \includegraphics[scale = 0.15]{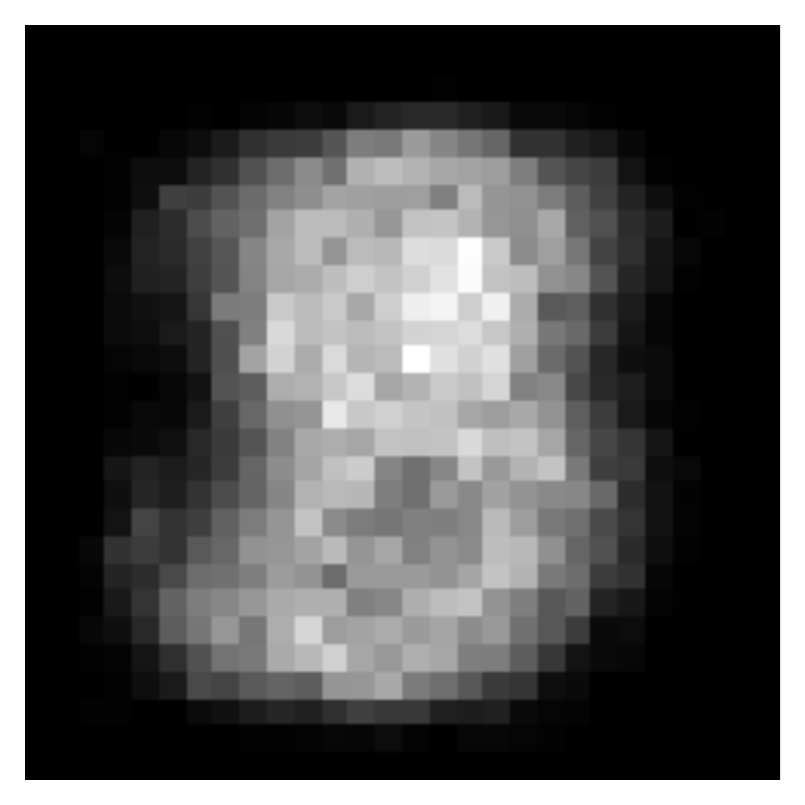}
    \end{minipage}
    \begin{minipage}{0.09\textwidth}
        \centering
        \includegraphics[scale = 0.15]{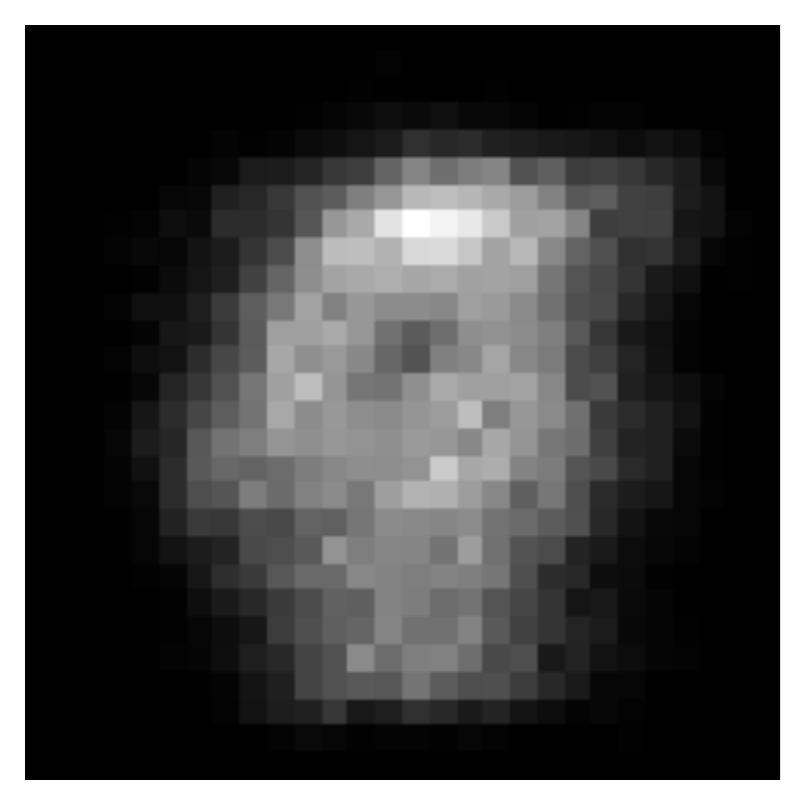}
    \end{minipage}
    \begin{minipage}{0.09\textwidth}
        \centering
        \includegraphics[scale = 0.15]{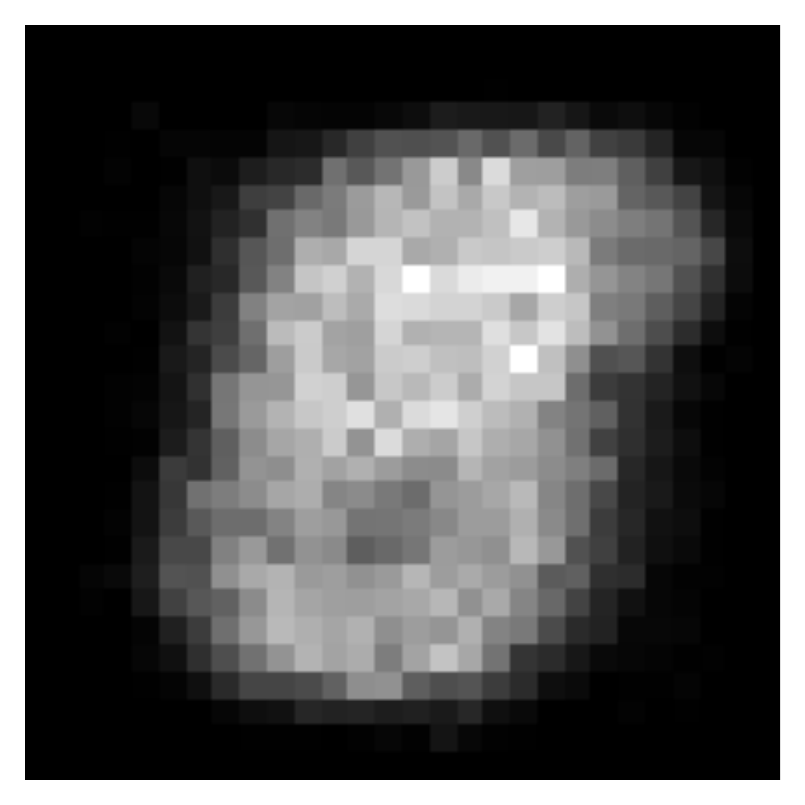}
    \end{minipage}
    \begin{minipage}{0.09\textwidth}
        \centering
        \includegraphics[scale = 0.15]{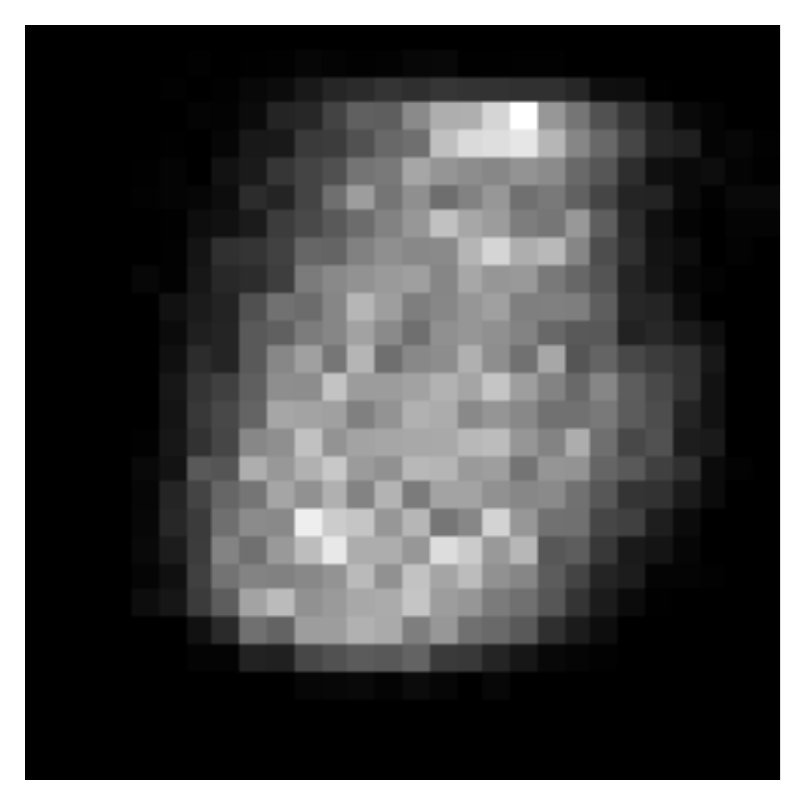}
    \end{minipage}
    \begin{minipage}{0.09\textwidth}
        \centering
        \includegraphics[scale = 0.15]{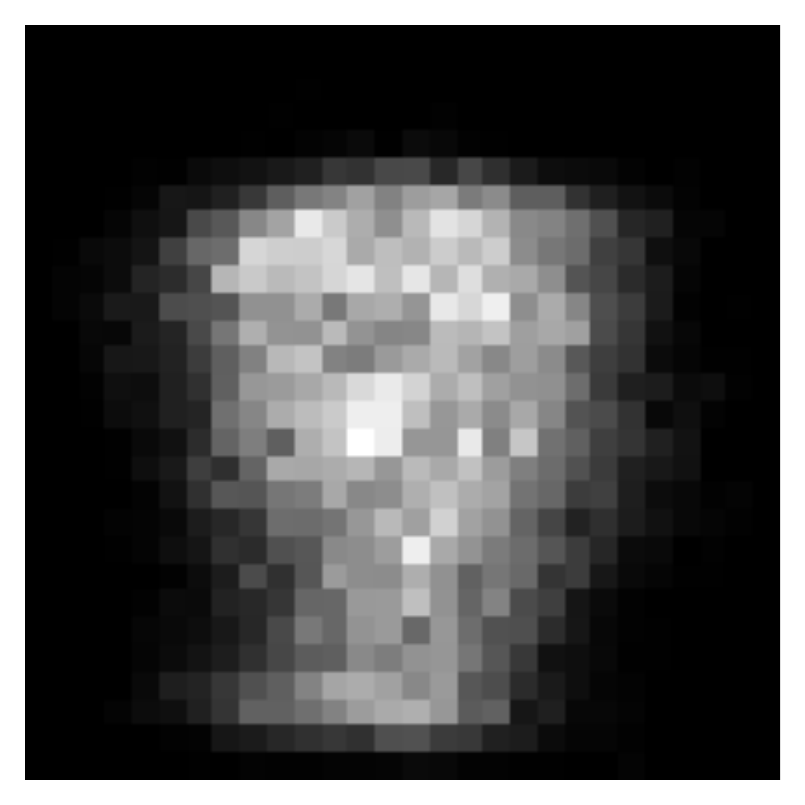}
    \end{minipage}
    \begin{minipage}{0.09\textwidth}
        \centering
        \includegraphics[scale = 0.15]{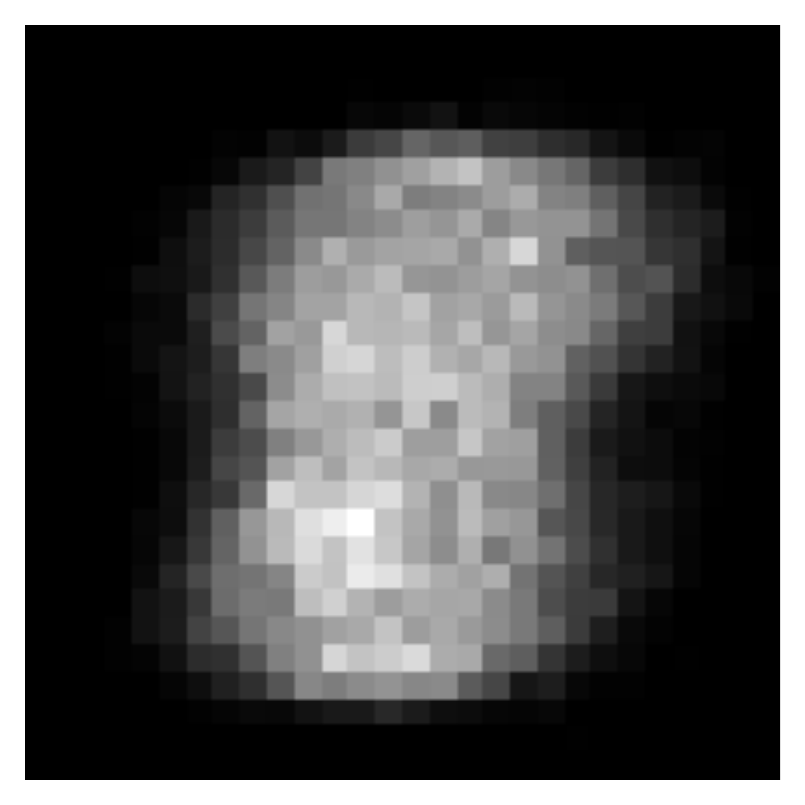}
    \end{minipage}
    \begin{minipage}{0.09\textwidth}
        \centering
        \includegraphics[scale = 0.15]{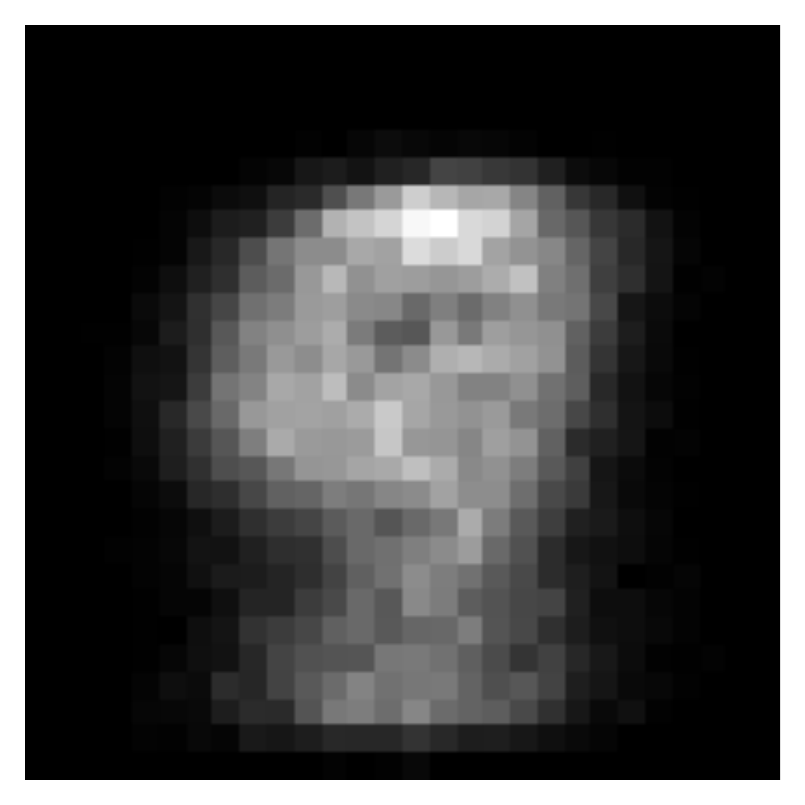}
    \end{minipage}
    
    \begin{minipage}{0.09\textwidth}
        \centering
        \includegraphics[scale = 0.15]{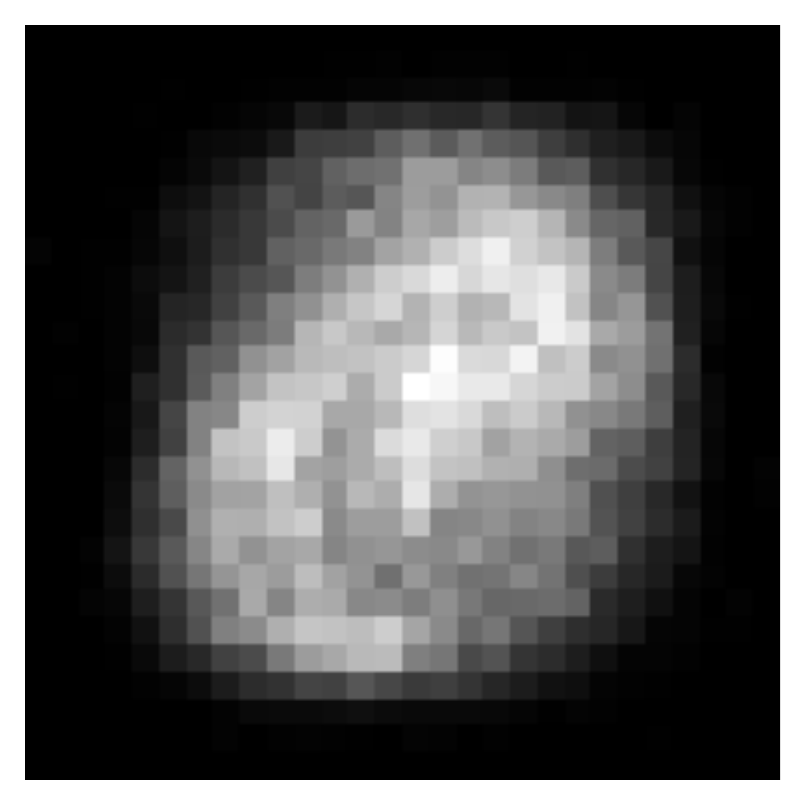}
    \end{minipage}
    \begin{minipage}{0.09\textwidth}
        \centering
        \includegraphics[scale = 0.15]{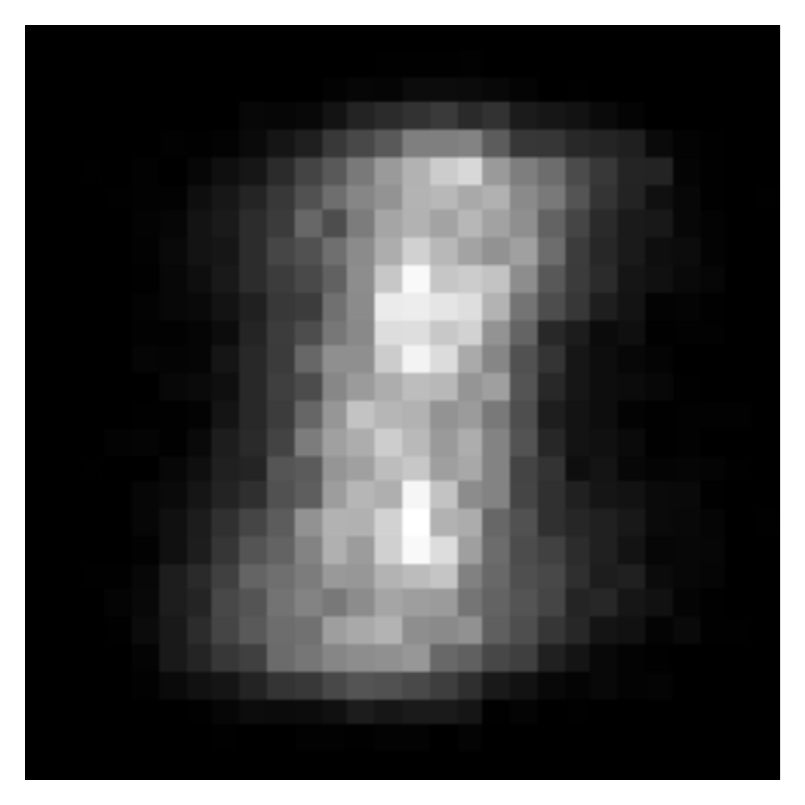}
    \end{minipage}
    \begin{minipage}{0.09\textwidth}
        \centering
        \includegraphics[scale = 0.15]{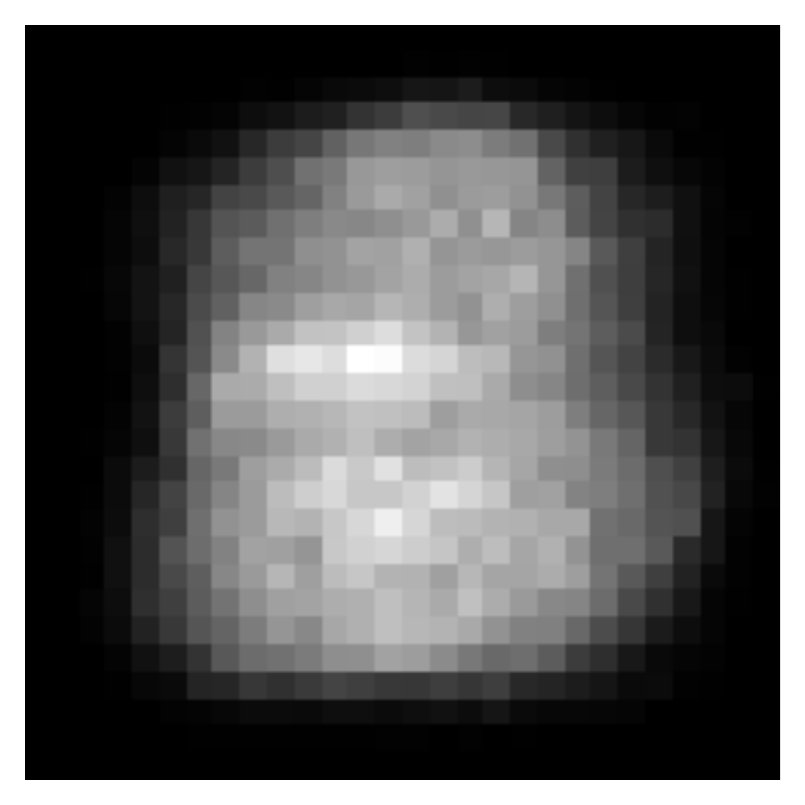}
    \end{minipage}
    \begin{minipage}{0.09\textwidth}
        \centering
        \includegraphics[scale = 0.15]{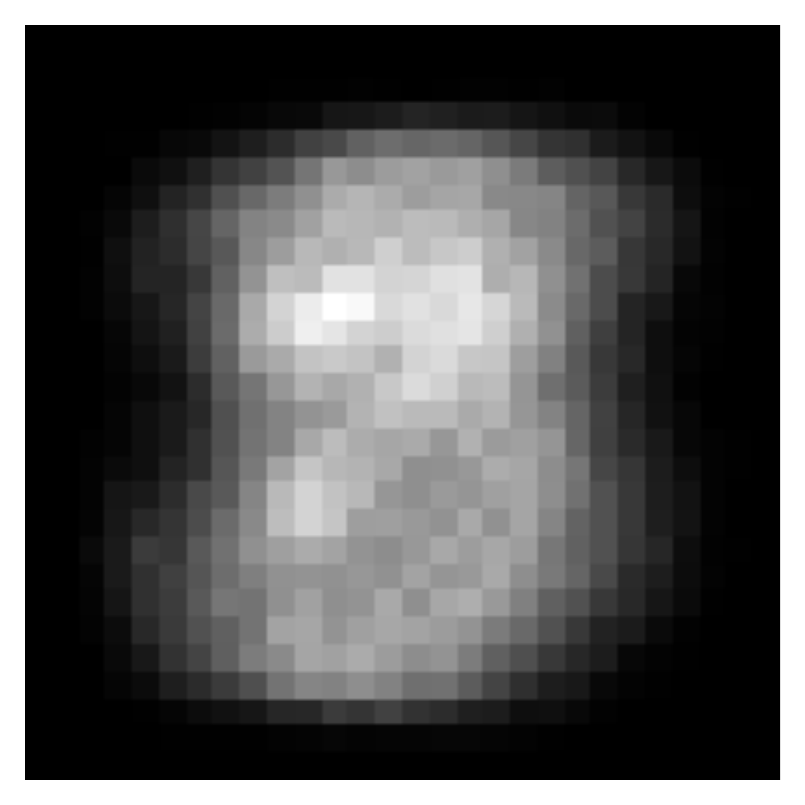}
    \end{minipage}
    \begin{minipage}{0.09\textwidth}
        \centering
        \includegraphics[scale = 0.15]{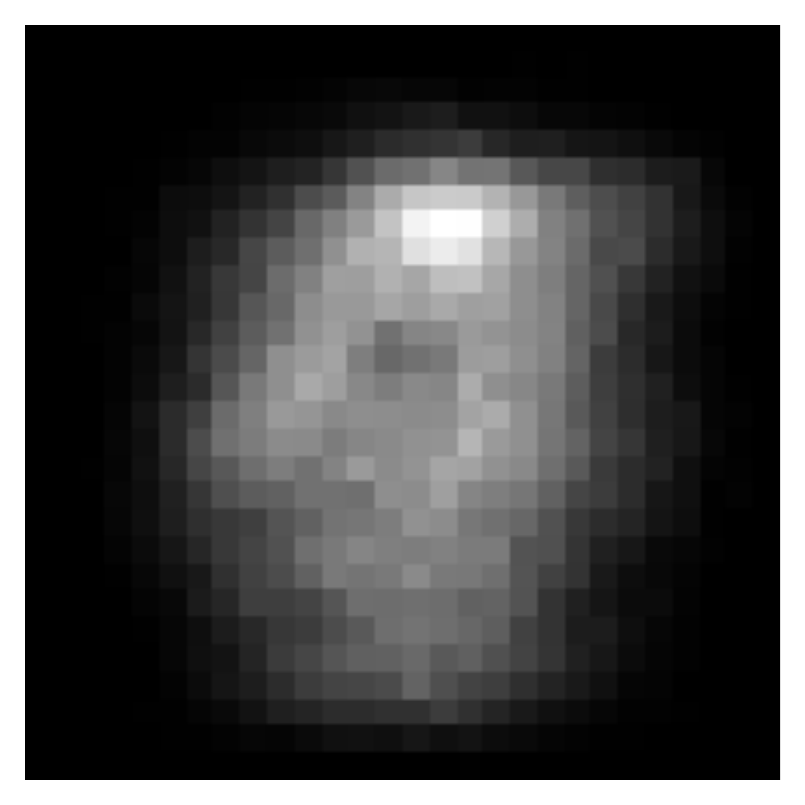}
    \end{minipage}
    \begin{minipage}{0.09\textwidth}
        \centering
        \includegraphics[scale = 0.15]{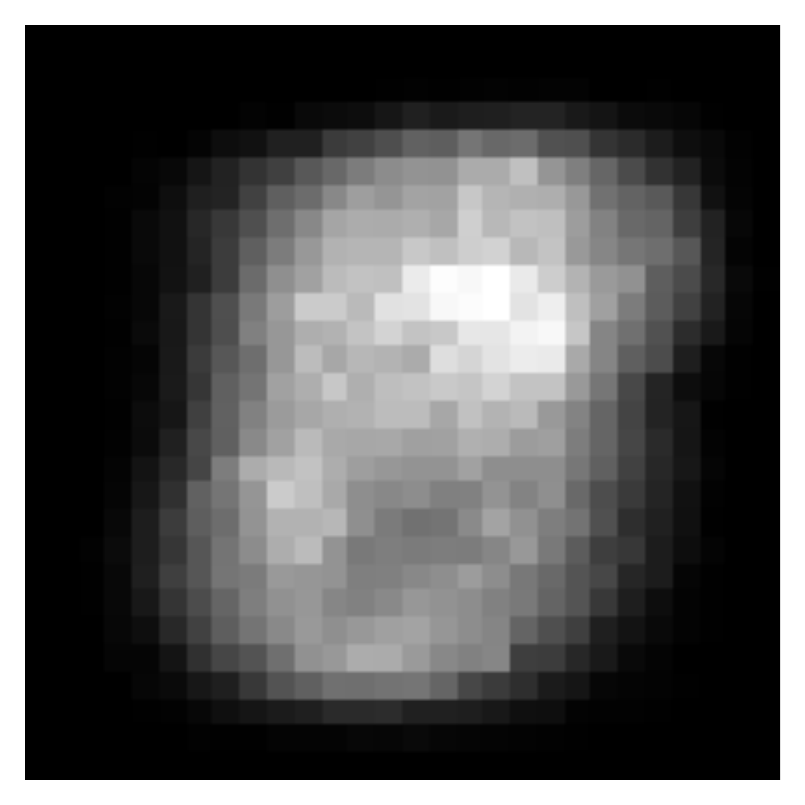}
    \end{minipage}
    \begin{minipage}{0.09\textwidth}
        \centering
        \includegraphics[scale = 0.15]{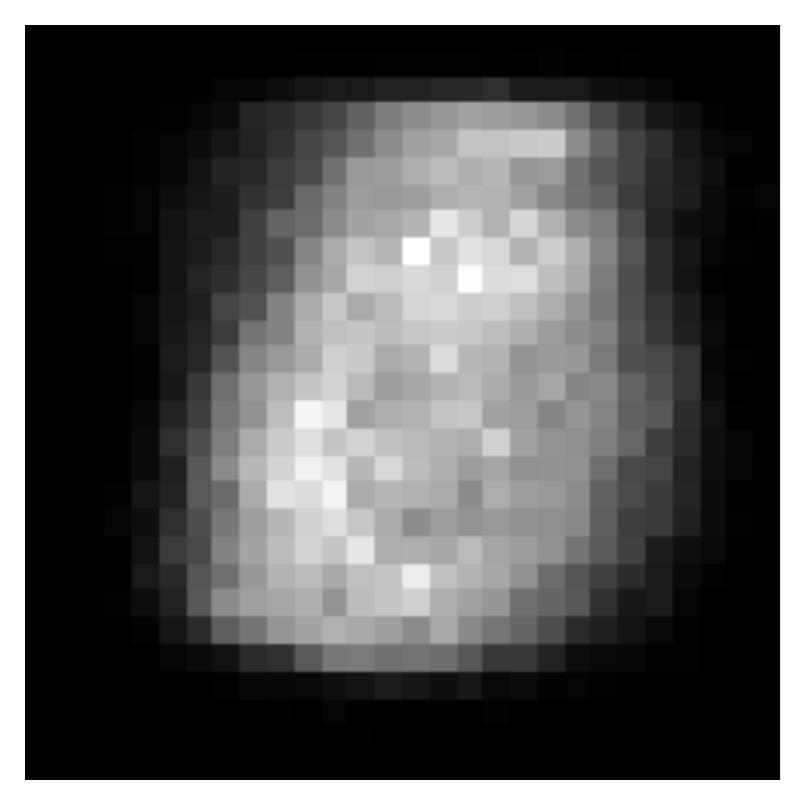}
    \end{minipage}
    \begin{minipage}{0.09\textwidth}
        \centering
        \includegraphics[scale = 0.15]{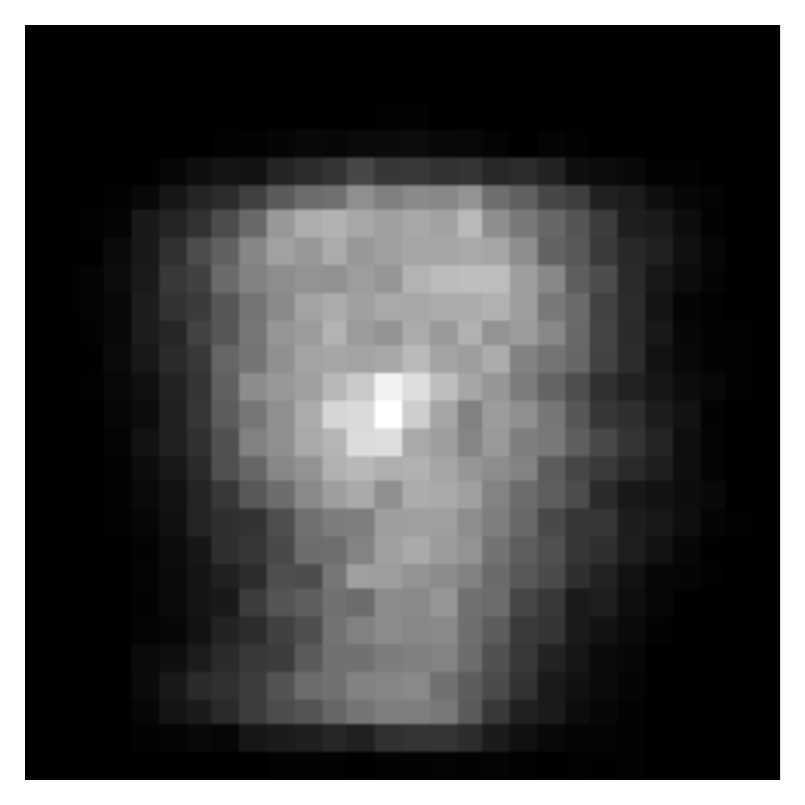}
    \end{minipage}
    \begin{minipage}{0.09\textwidth}
        \centering
        \includegraphics[scale = 0.15]{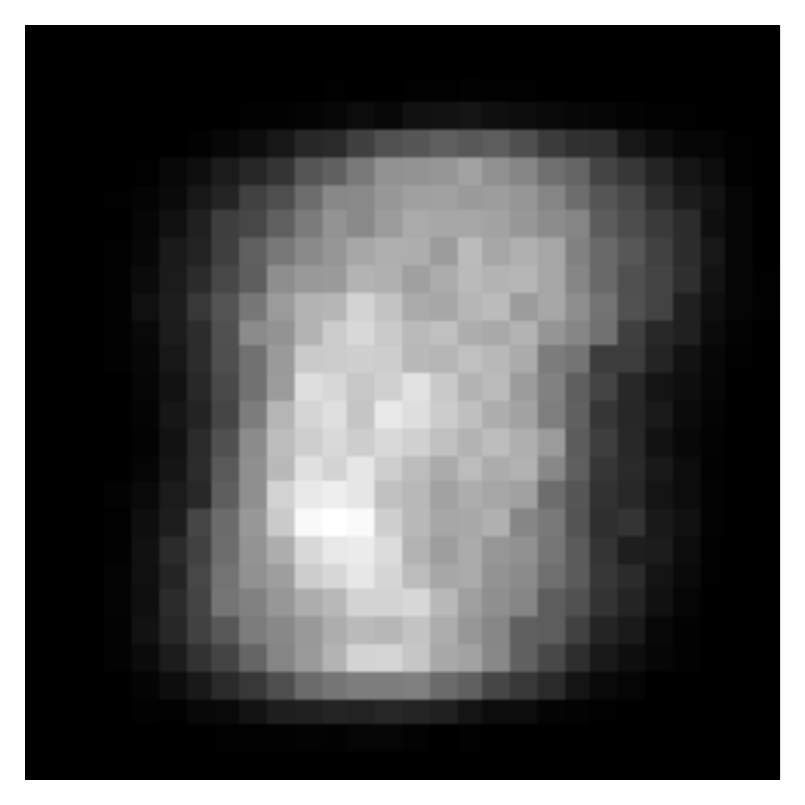}
    \end{minipage}
    \begin{minipage}{0.09\textwidth}
        \centering
        \includegraphics[scale = 0.15]{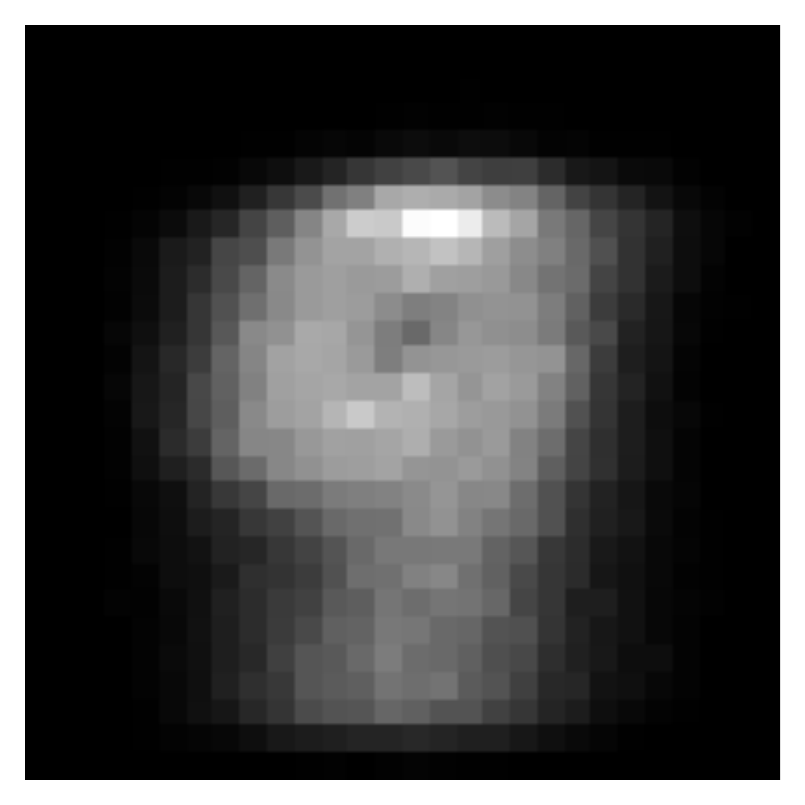}
    \end{minipage}
    \caption{Normalized risk (impurity) reduction importances for MNIST digits. Top: PU ET with nonnegative risk estimator and quadratic loss. Bottom: PN ET with quadratic loss.}
    \label{fig:normalised_importances}
\end{figure}

For MNIST, the plot of the importance scores for PU ET for each digit often suggests the shape of the digit. 
\Cref{fig:normalised_importances} shows that the normalized risk reduction importance makes many more pixels more important. This observation is consistent with our discussion in Section 4. Interestingly, the importances for PU ET appear to be very similar to those for PN ET using both the risk reduction importance and normalised risk reduction importance. This suggests that the learnt PU model is likely quite similar to the learnt PN model.

\begin{figure}[H]
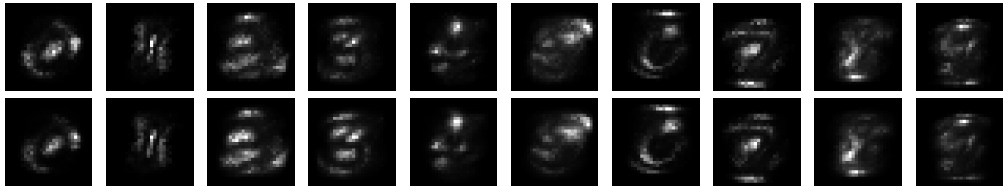

    \centering
    \begin{minipage}{0.09\textwidth}
        \centering
        \includegraphics[scale = 0.15]{figs/0nnPUunnormalised.pdf}
    \end{minipage}
    \begin{minipage}{0.09\textwidth}
        \centering
        \includegraphics[scale = 0.15]{figs/1nnPUunnormalised.pdf}
    \end{minipage}
    \begin{minipage}{0.09\textwidth}
        \centering
        \includegraphics[scale = 0.15]{figs/2nnPUunnormalised.pdf}
    \end{minipage}
    \begin{minipage}{0.09\textwidth}
        \centering
        \includegraphics[scale = 0.15]{figs/3nnPUunnormalised.pdf}
    \end{minipage}
    \begin{minipage}{0.09\textwidth}
        \centering
        \includegraphics[scale = 0.15]{figs/4nnPUunnormalised.pdf}
    \end{minipage}
    \begin{minipage}{0.09\textwidth}
        \centering
        \includegraphics[scale = 0.15]{figs/5nnPUunnormalised.pdf}
    \end{minipage}
    \begin{minipage}{0.09\textwidth}
        \centering
        \includegraphics[scale = 0.15]{figs/6nnPUunnormalised.pdf}
    \end{minipage}
    \begin{minipage}{0.09\textwidth}
        \centering
        \includegraphics[scale = 0.15]{figs/7nnPUunnormalised.pdf}
    \end{minipage}
    \begin{minipage}{0.09\textwidth}
        \centering
        \includegraphics[scale = 0.15]{figs/8nnPUunnormalised.pdf}
    \end{minipage}
    \begin{minipage}{0.09\textwidth}
        \centering
        \includegraphics[scale = 0.15]{figs/9nnPUunnormalised.pdf}
    \end{minipage}
    
    \begin{minipage}{0.09\textwidth}
        \centering
        \includegraphics[scale = 0.15]{figs/0PNunnormalised.pdf}
    \end{minipage}
    \begin{minipage}{0.09\textwidth}
        \centering
        \includegraphics[scale = 0.15]{figs/1PNunnormalised.pdf}
    \end{minipage}
    \begin{minipage}{0.09\textwidth}
        \centering
        \includegraphics[scale = 0.15]{figs/2PNunnormalised.pdf}
    \end{minipage}
    \begin{minipage}{0.09\textwidth}
        \centering
        \includegraphics[scale = 0.15]{figs/3PNunnormalised.pdf}
    \end{minipage}
    \begin{minipage}{0.09\textwidth}
        \centering
        \includegraphics[scale = 0.15]{figs/4PNunnormalised.pdf}
    \end{minipage}
    \begin{minipage}{0.09\textwidth}
        \centering
        \includegraphics[scale = 0.15]{figs/5PNunnormalised.pdf}
    \end{minipage}
    \begin{minipage}{0.09\textwidth}
        \centering
        \includegraphics[scale = 0.15]{figs/6PNunnormalised.pdf}
    \end{minipage}
    \begin{minipage}{0.09\textwidth}
        \centering
        \includegraphics[scale = 0.15]{figs/7PNunnormalised.pdf}
    \end{minipage}
    \begin{minipage}{0.09\textwidth}
        \centering
        \includegraphics[scale = 0.15]{figs/8PNunnormalised.pdf}
    \end{minipage}
    \begin{minipage}{0.09\textwidth}
        \centering
        \includegraphics[scale = 0.15]{figs/9PNunnormalised.pdf}
    \end{minipage}
    \caption{Risk reduction importance scores for MNIST digits. Top: PU ET with nonnegative risk estimator and quadratic loss. Bottom: PN ET with quadratic loss.}
\end{figure}

\section{Feature selection with risk reduction importance}
\begin{figure}[!h]
    \centering
    \includegraphics[scale = 0.7]{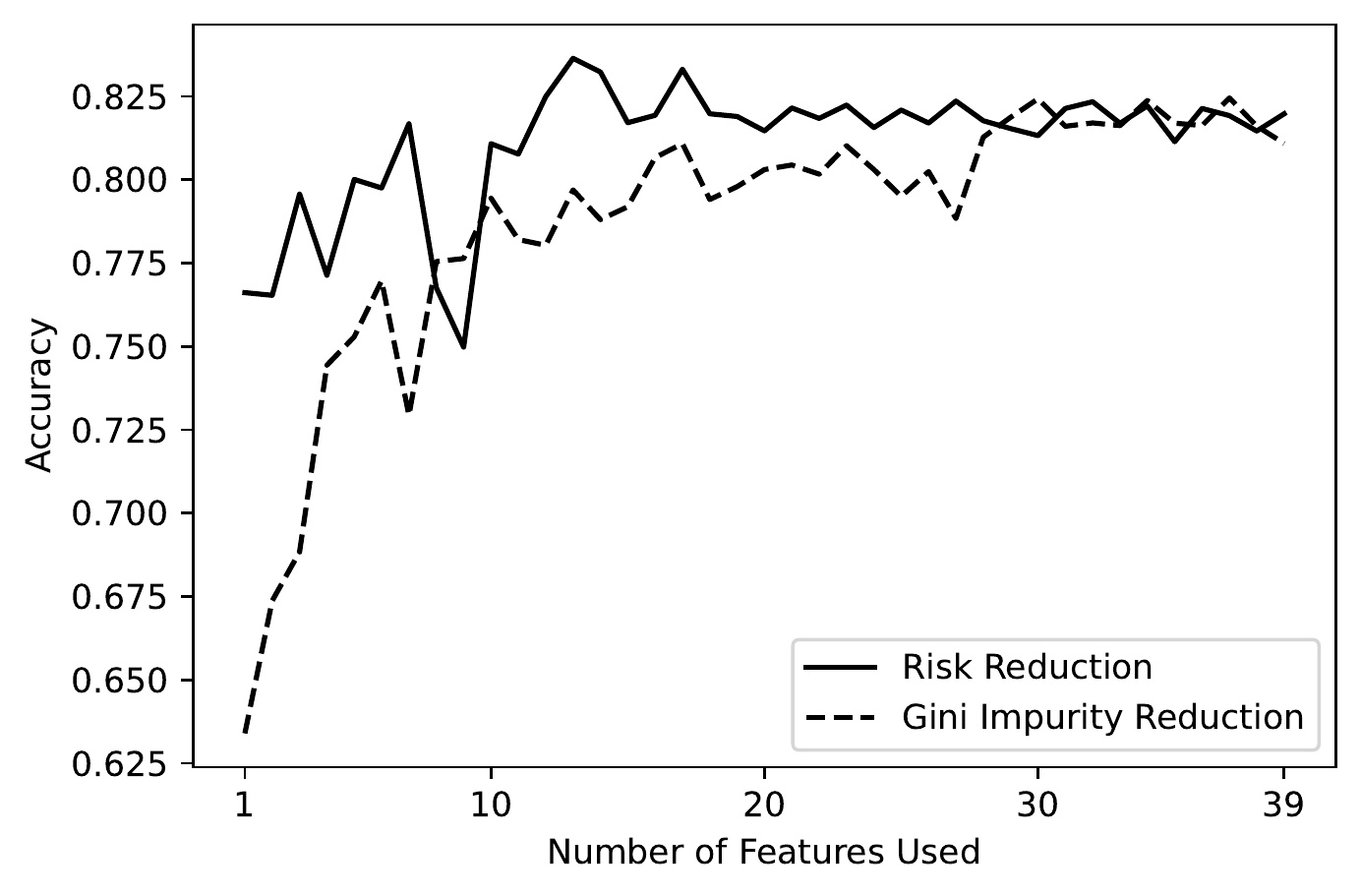}
    \caption{Risk reduction importance and Gini impurity reduction importance when using different numbers of features. }
    \label{fig:progressive}
\end{figure}

We demonstrate that the risk reduction importance is more capable of selecting
features which are responsible for risk minimization and the generalization
performance.
We do this on the UNSW-NB15 dataset.
\Cref{fig:progressive} shows how the accuracy of PU ET changes on UNSW-NB15 when
using only top features selected by risk reduction importance and Gini impurity
reduction importance. 
All accuracies are reported as the mean over five replications to account for
randomness in sampling the P data used during training and randomness in the
decision tree construction process. 

With the same number of selected features, the risk reduction importance generally leads to increased accuracy on the test data compared to using Gini impurity reduction on the UNSW-NB15 dataset. 
The difference in area under the curve is 0.84 in favour of the risk reduction importance. 
This confirms that the risk reduction importance is indeed more useful for
selecting features which are responsible for risk minimization and the generalization
performance.

% \pagebreak
%%%%%%%%%%%%%%%%%%%%%%%%%%%%%%%%%%%%%%%%%%%%%%%%%%%%%%%%%%%%%%%%%%%%%%%%%%%%%%%%%%%%%%%%%%%%%%%%%%%%%%%%

\end{document}